\documentclass[11pt]{article}

\usepackage[final]{acl}

\usepackage{times}
\usepackage{latexsym}

\usepackage[T1]{fontenc}

\usepackage[utf8]{inputenc}

\usepackage{microtype}

\usepackage{inconsolata}

\usepackage{graphicx}
\usepackage{amsmath}
\usepackage{amsfonts}
\usepackage{booktabs}
\usepackage{multirow} 
\usepackage{tikz}
\usetikzlibrary{arrows.meta, positioning, shapes.geometric, backgrounds}
\usepackage{listings}
\usepackage{xcolor}
\usepackage{tabularx}   
\usepackage{colortbl}   
\usepackage{adjustbox}  
\usepackage{subcaption}
\usepackage[table]{xcolor}
\usepackage{verbatim}
\usepackage{algorithm}
\usepackage{algorithmic}
\usepackage{array}
\usepackage{stfloats}
\usetikzlibrary{patterns, decorations.pathreplacing}

\definecolor{bestvalue}{HTML}{87CEEB} 
\definecolor{secondbest}{HTML}{E0FFFF} 
\title{BioProAgent: Neuro-Symbolic Grounding \\for Constrained Scientific Planning}

\author{%
  Yuyang Liu\textsuperscript{1,2,$\dagger$},
  Jingya Wang\textsuperscript{2}, Liuzhenghao Lv\textsuperscript{3}, Yonghong Tian\textsuperscript{1,2,3,$\dagger$}\\
  \textsuperscript{1}School of AI for Science, Peking University \\
\textsuperscript{2}School of Electronic and Computer Engineering, Peking University \\
\textsuperscript{3}School of Computer Science, Peking University \\
{\tt\small \{liuyuyang13, yhtian\}@pku.edu.cn, \{lvliuzh\}@stu.pku.edu.cn}
}

\begin{document}
\maketitle

\begin{abstract}
Large language models (LLMs) have demonstrated significant reasoning capabilities in scientific discovery but struggle to bridge the gap to physical execution in wet-labs. In these irreversible environments, probabilistic hallucinations are not merely incorrect; they can cause equipment damage or experimental failure.
We propose \textbf{BioProAgent}, a neuro-symbolic framework that anchors probabilistic planning in a deterministic Finite State Machine (FSM).
We introduce a State-Augmented Planning mechanism that enforces a rigorous \textit{Design-Verify-Rectify} workflow, ensuring hardware compliance before execution. Furthermore, we address the context bottleneck inherent in complex device schemas by \textit{Semantic Symbol Grounding}, reducing token consumption by $\sim$6$\times$ through symbolic abstraction. 
In the extended BioProBench benchmark, BioProAgent achieves 95.6\% physical compliance (compared to 21.0\% for ReAct), demonstrating that neuro-symbolic constraints are essential for reliable autonomy in irreversible physical environments. Code: \href{https://github.com/YuyangSunshine/bioproagent}{https://github.com/YuyangSunshine/bioproagent} | Website: \href{https://yuyangsunshine.github.io/BioPro-Project}{BioPro-Project}.
\end{abstract}

\begin{figure}[th]
    \centering
    \includegraphics[width=1.0\linewidth]{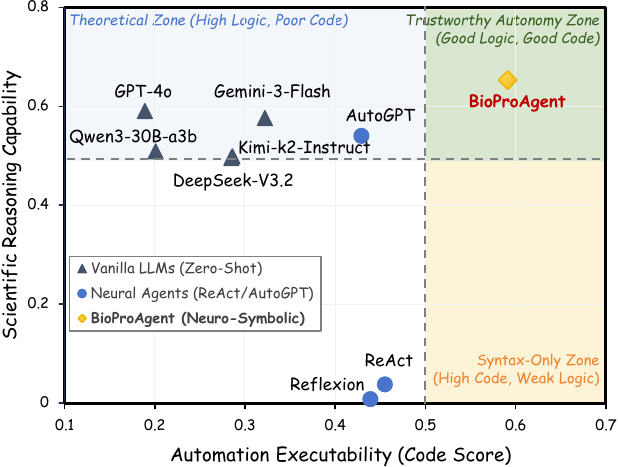} 
    \vspace{-20pt}
    \caption{\textit{Scientific Reasoning} vs. \textit{Automation Executability}. 
    \textbf{Vanilla LLMs} occupy the \textit{Theoretical Zone} (high logic, weak code), while \textbf{Neural Agents (like ReAct)} often generate better code but fail in scientific reasoning. 
    \textbf{BioProAgent} achieves \textit{Trustworthy Autonomy} with superior performance in both dimensions.}
    \label{fig:teaser_performance}
    \vspace{-10pt}
\end{figure}

\section{Introduction}~\label{sec:introduction}
Large Language Models (LLMs) are increasingly shifting from static knowledge retrieval to active world modeling in scientific discovery.
In the Self-Driving Laboratories (SDL) \citep{stein2019progress, abolhasani2023rise}, LLMs serve as central cognitive engines. LLMs, operating as Agentic AI \cite{sapkota2025ai, gridach2025agentic}, can design complex chemical synthesis schemes \cite{boiko2023autonomous}, discover novel materials \cite{ghafarollahi2024sciagents}, and act as laboratory assistants \cite{darvish2025organaroboticassistantautomated}.
Recently, multi-agent systems \citep{jin2025biolab,zhou2025prime} have demonstrated impressive capabilities in biological reasoning and computational workflow orchestration. 
However, a critical \textbf{Execution Gap} \cite{wang2024scibench} emerges when transitioning from computer simulation predictions to in vitro physical experiments.
Despite the remarkable physical autonomy achieved by robotic platforms, from mobile robotic chemists \cite{burger2020mobile} to autonomous materials synthesis labs \cite{szymanski2023autonomous}, they still operate under rigid, task-specific workflows.
Unlike reversible code and virtual environments in software engineering, due to irreversible physical laws, any deviation in parameters can lead to catastrophic equipment failures, sample loss, or unreproducible experimental in biological experimental environments.

Despite the reasoning capabilities of existing scientific agents, they still face key challenges in high-risk automated applications.
First, standard agents \cite{yao2022react,shinn2023reflexion} suffer from Cognitive Drift, where contextual overload leads to process confusion, resulting in skipped critical steps or confused temporal dependencies. 
While current memory systems \cite{chhikara2025mem0, packer2023memgpt} excel at semantic retrieval, they struggle with lossless tracking of distinct physical entities. A slight ``lost-in-the-middle'' \cite{liu2024lost} confusion between similar reagent IDs can cause experimental failure.
More critically, there is a distinct lack of Pre-Execution Interlocks. While previous neuro-symbolic methods like PAL \cite{gao2023pal}, Program-of-Thoughts \cite{chen2022program}, and SayCan \cite{ahn2022can} have combined LLMs with planners to improve logical correctness, they primarily focus on reasoning optimality. They cannot handle the strict and irreversible safety constraints in wet labs because they lack deterministic control mechanisms based on physical rules that stop and self-correct before execution.

To address these challenges, we propose a neuro-symbolic \textbf{BioProAgent} for achieving trustworthy autonomy (Fig.~\ref{fig:teaser_performance}), which grounds probabilistic reasoning in a deterministic backbone network with a training-free \textbf{Finite State Machine (FSM)}.
Unlike static agents restricted to pre-defined workflows, BioProAgent flexibly adapts to diverse tasks by using FSM as a safety boundary to enforce a rigorous ``Design-Verify-Rectify'' workflow.
This controller ensures that all hardware instructions must undergo hierarchical verification for both scientific logic and physical safety before being issued. Furthermore, to address context challenges, we introduce \textbf{Semantic Symbol Grounding}. By decoupling the high-dimensional payloads into symbolic pointers, we reduce token consumption by $\sim$6$\times$ while ensuring 100\% resource consistency. Extensive evaluation on extended BioProBench \cite{bioprotocolbench2025} shows that BioProAgent achieves an 90.0\% success rate in error recovery, compared to a complete failure (0\%) for the standard baseline.

Our contributions are summarized as follows:
\begin{itemize}
    \item We propose BioProAgent, a training-free neuro-symbolic framework that bridges the gap between probabilistic reasoning and deterministic physical execution.
    \item We introduce a State-Augmented Planning mechanism driven by a deterministic FSM and a Semantic Symbol Grounding technique that effectively addresses context problem.
    \item Extensive experiments demonstrate BioProAgent achieves state-of-the-art performance in scientific validity and physical consistency, and significantly outperforms baselines.
\end{itemize}

\section{Related Works}~\label{sec:relatedworks}
\vspace{-10pt}
\subsection{LLM Agents for Scientific Discovery}
Autonomous research has evolved from basic tool usage \citep{bran2023chemcrow,boiko2023autonomous} to complex multi-agent collaboration \citep{zhang2025multimodal,jin2025biolab,huang2025biomni}. 
Recent frameworks like {DeepScientist} \cite{weng2025deepscientist} and {Organa} \cite{darvish2025organaroboticassistantautomated} demonstrate long-horizon autonomy, and {BioMARS} \cite{qiu2025biomars} and {AutoLabs} \cite{panapitiya2025autolabs} introduce domain-specific visual monitoring and self-correction. AI-native operating systems \cite{sim2024chemos, fei2024alabos, gao2025unilabos} are also continuously strengthening the execution infrastructure. However, despite these advances, a critical gap remains: the lack of cognitive safety interlocking mechanisms, making general-purpose agents vulnerable to irreversible physical damage.

\subsection{Neuro-Symbolic Reasoning \& Action}
Coupling LLMs with symbolic planners or code interpreters, such as {PAL} \cite{gao2023pal}, {Program-of-Thoughts} \cite{chen2022program}, and {LLM+P} \cite{liu2023llmp}, significantly boosts logical and planning accuracy. In embodied settings, {SayCan} \cite{ahn2022can} and {Voyager} \cite{wang2023voyager} ground actions in physical affordances, while iterative frameworks like {ReAct} \cite{yao2022react} and {Reflexion} \cite{shinn2023reflexion} improve reasoning via verbal feedback. Recent work~\cite{hyun2025aligning} utilizes physics-aware rejection sampling to align LLM reasoning traces for materials discovery. While these methods are effective in completing tasks in reversible or simulated environments, they lack mechanisms for enforcing irreversible safety constraints.

\subsection{Long-Horizon Context Management}
While Longformer \cite{beltagy2020longformer} and RAG \cite{lewis2020retrieval} extend context windows, they often fragment the coherent narrative essential for protocol execution. Agent-specific memory systems like MemGPT \cite{packer2023memgpt} and Mem0 \cite{chhikara2025mem0} maintain multi-session coherence through hierarchical compression. Recent approaches such as ACON \cite{kang2025acon} and LLM-State \cite{chen2024llmstate} address long-horizon tasks via learned state compression. However, these purely neural methods remain probabilistic and inherently lossy, making them vulnerable to ``lost-in-the-middle'' phenomenon \cite{liu2024lost}. This ambiguity is unacceptable for tracking precise scientific entities like reagent IDs. 

\begin{figure*}[t]
    \centering
    \includegraphics[width=\linewidth]{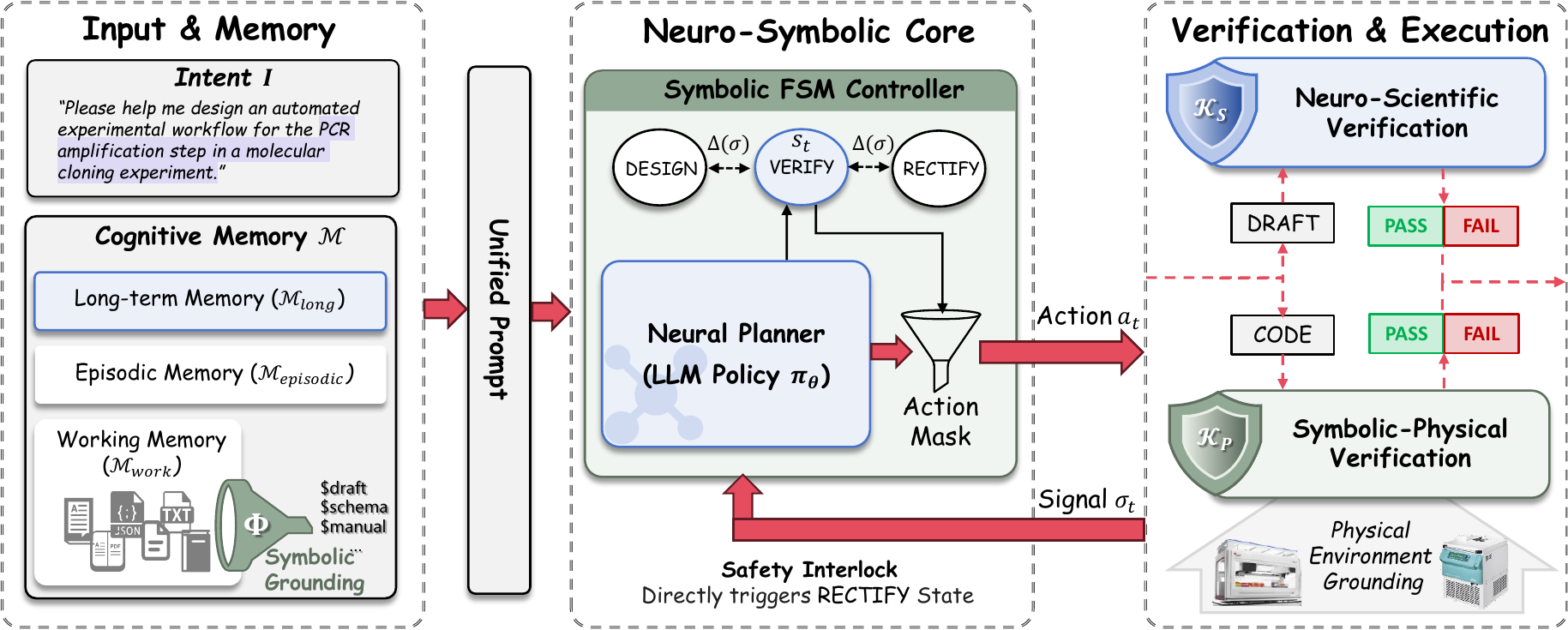}
    \vspace{-15pt}
    \caption{\textbf{Overview of BioProAgent.} (1) Cognitive Memory utilizes Symbolic Grounding $\Phi$ to manage context efficiently; (2) Neural Planner $\pi_\theta$ is grounded in a Design-Verify-Rectify FSM $\Delta(\sigma)$; (3) Hierarchical Verification ($\mathcal{K}_s, \mathcal{K}_p$) acts as a safety interlock, enforcing physical compliance by deterministically triggering rectification.}
    \label{fig:architecture}
\end{figure*}

\section{Methodology}~\label{sec:method}
\vspace{-15pt}

We formulate automated biological protocol generation as a constrained scientific planning problem.
Let $\mathcal{P}$ denote possible protocol sequences derived from the valid action space $\mathcal{A}$ defined by the Hardware Registry $\Omega$.
Given an intent $I$ and environmental context $\mathcal{E}$, our goal is to identify the optimal executable protocol $P^* \in \mathcal{P}$ that is scientifically valid and physically executable, as:
\begin{equation}
\resizebox{0.89\hsize}{!}{$
P^* = \underset{P \in \mathcal{P}}{\text{argmax}} \; \mathbb{P}_{\theta}(P|I,\mathcal{E}) \cdot \mathbb{I}[\mathcal{K}_s(P)] \cdot \mathbb{I}[\mathcal{K}_p(P)],
$}
\label{eq:objective}
\end{equation}
where $\mathbb{I}[\cdot]$ is indicator function. 
$\mathcal{K}_s$ and $\mathcal{K}_p$ the Scientific and Physical verification functions, respectively. (Summary of notations in Appendix~\ref{sec:appendix_notations}.) As illustrated in Fig.~\ref{fig:architecture}, our proposed BioProAgent consists of three main modules: a Hybrid Cognitive Memory, a Neuro-Symbolic Core driven by a deterministic FSM, and a Hierarchical Verification framework.

\subsection{Hybrid Cognitive Memory Architecture}
\label{sec:memory}
Existing agents often suffer from cognitive drift in long-horizon tasks, where early context regarding reagent IDs is lost.
To address this, we design a hierarchical memory system $\mathcal{M}=\langle\mathcal{M}_{work},\mathcal{M}_{episodic},\mathcal{M}_{long}\rangle$, which decouples reasoning logic from high-dimensional data payloads.

\paragraph{Structured Working Memory $\mathcal{M}_{work}$.} Direct input of raw JSON schemas (often >10k tokens) induces context saturation. We propose \textbf{Semantic Symbol Grounding}, which projects complex data artifacts into symbolic references with lightweight semantic previews. During planning, the working context is compressed via a projection function $\Phi$:
\begin{equation}
    \resizebox{0.89\hsize}{!}{$
\begin{split}
C_{context} &\leftarrow \Phi(\mathcal{M}_{work})\\ &= \{(k_i, \text{preview}(v_i))  \mid (k_i, v_i) \in \mathcal{M}_{work}\}.
\end{split}
$}
\end{equation}

For execution, a generated symbolic action $a_t$ is grounded via a resolution function $\Psi$: $a_t^{exec} \leftarrow \Psi(a_t, \mathcal{M}_{work})$, which dynamically maps pointers $k$ back to their complete physical parameters $v$. This reduces processing complexity from $\mathcal{O}(|v|)$ to $\mathcal{O}(j)$ where $j \ll |v|$ is the dimension of the symbolic pointer, ensuring the planner focuses on logic flow rather than data parsing.

\paragraph{Episodic and Long-term Memory.} 
To prevent process disorientation, the episodic memory $\mathcal{M}_{episodic}$ tracks the active FSM-guided trajectory $\tau_t=[(s_0, a_0, \sigma_0), \dots, (s_{t-1}, a_{t-1}, \sigma_{t-1})]$, which explicitly records the cognitive states, executed actions, and returned constraint signals. Meanwhile, the long-term memory $\mathcal{M}_{long}$ persists cross-session domain knowledge. Rather than statically bloating the prompt, $\mathcal{M}_{long}$ is dynamically queried via retrieval tools on demand. 

Finally, the neural planner synthesizes these active layers to form a unified contextual prompt: $Prompt_t = \text{Retrieved}(\mathcal{M}_{long}) \oplus \tau_t \oplus \Phi(\mathcal{M}_{work}) \oplus s_t$, ensuring procedural awareness without context saturation.



\subsection{State-Augmented Neuro-Symbolic Plan}
\label{sec:planning}
To address the lack of strict adherence to state-related rules required by safety-critical hardware in probabilistic LLMs, BioProAgent uses a deterministic finite state machine (FSM) to decouple reasoning from control, aiming to enforce a rigorous Design-Verify-Rectify (DVR) workflow.
The system is formally defined as $FSM=\langle \mathcal{S},\Sigma,\mathcal{A},\Delta,\pi_\theta\rangle$. Unlike finite state automata used in traditional robots, which hard-code specific operation sequences (e.g., \texttt{Move\_A\_to\_B}), our states (e.g., \texttt{DESIGN}, \texttt{VERIFY}, \texttt{RECTIFY}) control the cognitive processes of generation and verification, as illustrated in Fig.~\ref{fig:architecture}.
This abstraction allows BioProAgent to maintain a consistent control architecture across different domains, from molecular cloning to chemical synthesis, without requiring architectural reconstruction.

\begin{algorithm}[t]
\small
\renewcommand{\arraystretch}{1.2}
\caption{BioProAgent FSM-Gated Execution}
\label{alg:bioproagent}
\begin{algorithmic}[1]
\REQUIRE Intent $I$, Hardware Registry $\Omega$, Knowledge $\mathcal{M}_{long}$
\ENSURE Protocol $P^*$ or Failure
\STATE \textbf{Initialize:} $\mathcal{M}_{work} \leftarrow \text{Parse}(I)$, $s_0 \leftarrow \text{INIT}$
\STATE \textbf{Initialize:} Trajectory $\tau \leftarrow []$, Signal $\sigma \leftarrow \mathbf{0}$, $t \leftarrow 0$
\WHILE{$s_t \neq \text{SUCCESS}$ \textbf{and} $t < T_{max}$}
    \STATE $s_{t} \leftarrow \Delta(\sigma, s_{t-1})$ \COMMENT{Refer to Priority Matrix (Tab.~\ref{tab:decision_logic})}
    
    \IF{$s_t \in \{\text{RECTIFY\_DRAFT}, \text{RECTIFY\_CODE}\}$}
        \STATE Update prompt via $\sigma$
    \ENDIF
    
    \STATE Generate $a_t \sim \pi_{\theta}(a \mid s_t, \Phi(\mathcal{M}_{work}), \tau)$
    
    \IF{$a_t$ is \texttt{Draft}}
        \STATE Check: $\mathcal{K}_s(a_t)$; Update: $\sigma_{sci} \in \{-1, 1\}$
    \ELSIF{$a_t$ is \texttt{Code}}
        \STATE Check: $\mathcal{K}_p(a_t, \Omega)$; Update: $\sigma_{phys} \in \{-1, 1\}$
        \IF{$\sigma_{phys} = -1$}
            \STATE $\sigma \leftarrow \text{INTERLOCK}$ \COMMENT{Force Transition via Matrix}
        \ELSE
            \STATE Append valid action to $P^*$: $P^* \leftarrow P^* \cup \{a_t\}$
        \ENDIF
    \ENDIF
    
    \STATE Update Memory $\mathcal{M}_{episodic}$: $\tau \leftarrow \tau \cup \{(s_t, a_t, \sigma)\}$
    \STATE $t \leftarrow t + 1$
\ENDWHILE
\RETURN $P^*$
\end{algorithmic}
\end{algorithm}

\subsubsection{Neuro-Symbolic Control Engine}
To capture both resource availability and validation outcomes, we define the context signal space as $\Sigma = \{0,1\}^3 \times \{-1,0,1\}^2$. Let the signal vector at time $t$ be $\sigma_t = [\sigma_{know}, \sigma_{draft}, \sigma_{code}, \sigma_{sci}, \sigma_{phys}] \in \Sigma$. 
Here, resource signals take boolean values $\{0, 1\}$, while validation signals take ternary values $\{-1, 0, 1\}$ (representing \textit{Failed}, \textit{Pending}, and \textit{Passed}, respectively).

State transitions $\Delta: \mathcal{S} \times \Sigma \rightarrow \mathcal{S}$ are governed by a deterministic \textit{Priority Decision Matrix} (Table~\ref{tab:decision_logic}). 
The physical execution function $E_{phys}$ for an action $a_t$ is defined as:
\begin{equation}
\label{eq:interlock}
\resizebox{0.89\hsize}{!}{$
E_{phys}(a_t) =
\begin{cases}
\text{Execute}(a_t), & \text{if } a_t \text{ is Code} \land (\sigma_{phys} = 1) \\
\emptyset, & \text{otherwise (Interlock Triggered)}
\end{cases}
$}
\end{equation}

We formalize the system's reliability as a conditional safety guarantee:
$P(\text{Unsafe} \mid \text{Interlock}, \Omega, \Phi) \approx 0$. 
$\text{Interlock}$ denotes the FSM gating mechanism. The guarantee holds under two assumptions: (1) the hardware registry $\Omega$ completely covers the physical constraints; and (2) the semantic symbol grounding $\Phi$ correctly maps natural language intents to specific device IDs.

For unmodeled risks outside $\Omega$, the system conservatively aborts execution and returns a \texttt{FAILURE} signal to ensure hardware safety.
The FSM evaluates the ordering rules and selects the target state based on the highest-priority signal match:
\begin{equation}
s_{t+1} = \Delta(\sigma_t, s_t) = \text{Target}(r_j) \in \mathcal{S}
\end{equation}
where $r_j$ represents the highest-priority matched rule in Table~\ref{tab:decision_logic}, determined by $j = \min \{i \mid \text{Match}(r_i, \sigma_t)\}$. If technical validation fails, the interlock strictly prohibits execution. The complete neuro-symbolic execution, coordinating the interaction between $\Delta$ and $\pi_{\theta}$, is formalized in Algorithm~\ref{alg:bioproagent}.

\begin{table}[t]
\centering
\small
\renewcommand{\arraystretch}{1.2}
\caption{\textbf{FSM Priority Decision Matrix.} See Appendix~\ref{app:fsm} for the complete definition.}
\label{tab:decision_logic}
\vspace{-5pt}
\resizebox{\linewidth}{!}{%
\begin{tabular}{lcc}
\toprule
\textbf{Signal Condition} ($\sigma_t$) & \textbf{Priority} & Target State ($s_{t+1}$)\\
\midrule
$(\sigma_{phys} = -1) \land \sigma_{code}$ & 1 & \cellcolor{red!10}RECTIFY\_CODE  \\
$(\sigma_{sci} = -1) \land \sigma_{draft}$ & 2 & RECTIFY\_DRAFT  \\
$\neg (\sigma_{draft}) \land \sigma_{know}$ & 3 & DESIGN\_DRAFT  \\
$\sigma_{draft} \land (\sigma_{sci} = 1)$ & 4 & DESIGN\_CODE  \\
$\sigma_{draft} \land (\sigma_{sci} = 0)$ & 5 & VERIFY\_DRAFT  \\
$\sigma_{code} \land (\sigma_{phys} = 1)$ & 6 & SUCCESS \\
... & ... & ... \\
\bottomrule
\end{tabular}%
}
\end{table}

\subsubsection{Neural Reasoning Policy}
In each state $s_t$, $\pi_\theta$ selects an action $a_t$ from $\mathcal{A}$, corresponding to the set of available tools (Appendix~\ref{app:tools}) categorized into three functional clusters: (1) \textbf{Design} (Generation), (2) \textbf{Verify} (Validation), and (3) \textbf{Rectify} (Correction). 
To ensure coherence, the FSM dynamically prunes the action space to $\mathcal{A}_{valid} \subset \mathcal{A}$ based on the active phase in $s_t$. 
The planner is conditioned on $s_t$, and the system prompt explicitly enforces this state-dependent action masking (Appendix~\ref{app:prompt_planner}).

\subsection{Hierarchical Verification Framework}
\label{sec:verification}
To implement the \textbf{Verify} phase of the DVR loop, we employ a two-layer hierarchical framework to enforce the constraints defined in Eq.~(\ref{eq:objective}). Upon execution, the binary outputs of these indicator functions are mapped directly to our FSM's ternary validation signals ($\sigma_{sci}$ and $\sigma_{phys}$), where an indicator value of $1$ yields a \textit{Passed} signal ($1$) and $0$ yields a \textit{Failed} signal ($-1$).

\paragraph{Layer 1: Neuro-Scientific Verification ($\mathcal{K}_s$).}
The Scientific Reflector assesses the \textbf{Design} draft using prompts derived from expert-in-the-loop iterations. This ensures the CoT reasoning~\cite{wei2022chain} follows professional wet-lab standards rather than superficial syntax checking. We formalize this evaluation as:
\begin{equation}
\resizebox{0.89\hsize}{!}{$
\mathbb{I}[\mathcal{K}_{s}(P_{draft})] = \mathbb{I} \left[ \text{Reflector}(P_{draft}, \text{Task}) = \text{``PASS''} \right].
$}
\end{equation}
Full verification criteria, including standard controls and biosafety checks, are detailed in Appendix~\ref{app:prompt_alignment}.

\paragraph{Layer 2: Symbolic-Physical Verification ($\mathcal{K}_{p}$).}
To ensure safety, a deterministic \textit{Rule Engine} ($\mathcal{R}_{phys}$) verifies the generated machine code against the hardware registry $\Omega$. The registry $\Omega$ encodes constraints for 22 instruments across Liquid Handling (LH), Thermal Control (TC), and Centrifugation (CF). 
Crucially, any parameter violation drives the overall product to zero (i.e., $\mathbb{I}[\mathcal{K}_p] = 0$, triggering $\sigma_{phys} = -1$), which deterministically activates the \textbf{Rectify} interlock:
\begin{equation}
\resizebox{0.89\hsize}{!}{$
\mathbb{I}[\mathcal{K}_{p}(P_{code})] = \prod_{op \in P_{code}} \prod_{c \in \Omega(\text{device}_{op})} \mathbb{I}[\text{Check}(op, c)],
$}
\end{equation}
where $op$ is an individual atomic operation within the generated sequence $P_{code}$, $\text{device}_{op}$ denotes the specific hardware instrument targeted by $op$, and $c$ represents an individual physical constraint (e.g., maximum speed, temperature limits) defined in the registry $\Omega$ for that device. The Boolean function $\text{Check}(op, c)$ returns $1$ if the operation complies with the constraint, and $0$ otherwise.

\section{Experiments}~\label{sec:experiments}
\vspace{-10pt}
\subsection{Experimental Setup}
\paragraph{Benchmark and Hardware Grounding.}
We evaluate our framework using an extended version of {BioProBench}~\cite{bioprotocolbench2025}, including four specialized subsets:
\textbf{Subset A (Protocol Drafting)} assesses scientific validity;
\textbf{Subset B (Code Generation)} evaluates hardware schema compliance and parameter accuracy;
\textbf{Subset C (Long-Horizon)} targets global orchestration, featuring 9 complex protocols with 547 atomic steps (max 238 per task);
and \textbf{Subset D (Error Correction)} measures robustness against injected faults.
To ensure reproducibility, the complete source code, the digitized hardware registry ($\Omega$) covering 22 core synthetic biology instruments, and the dataset are provided in the supplementary material.
Crucially, although evaluated in simulation, all generated instructions strictly adhere to the API specifications of standard automated devices to bridge the sim-to-real gap.

\paragraph{Baselines.}
We compare BioProAgent against three categories of baselines:
(1) \textbf{Vanilla LLMs}: GPT-4o~\citep{gpt-4o}, Gemini-3-Flash~\cite{gemini3flash2025}, DeepSeek-V3.2~\cite{deepseekv3.2}, and Kimi-k2-instruct~\cite{kimi-k2} via direct prompting;
(2) \textbf{Standard Agents}: ReAct~\cite{yao2022react}, Reflexion~\cite{shinn2023reflexion}, and AutoGPT~\cite{yang2023auto};
(3) \textbf{Domain-Specific Agent}: Biomni~\cite{huang2025biomni} (evaluated on native protocol generation).
\noindent\textit{Critical Implementation Detail:} To ensure fair comparison, all standard agent baselines were equipped with the {exact same toolset} as BioProAgent. 

\begin{table}[t]
\centering
\small
\renewcommand{\arraystretch}{1.2}
\caption{\textbf{The BioProAgent Evaluation Framework.} Metrics are categorized by dimension, improvement direction ($\uparrow$/$\downarrow$), and evaluation source.}
\label{tab:evaluation_metrics}
\vspace{-5pt}
\resizebox{\linewidth}{!}{
\begin{tabular}{@{}l l l@{}}
\toprule
\textbf{Dimension} & \textbf{Evaluator} & \textbf{Key Metrics} \\ 
\midrule
\multirow{2}{*}{\textbf{Scientific}} 
 & Semantic Match & $S_{sem}$ ($\uparrow$), ROUGE-L ($\uparrow$) \\
 & LLM-as-a-Judge & $C_{s}$ ($\uparrow$) \\ 
\cmidrule{1-3}
\multirow{2}{*}{\textbf{Physical}} 
 & Rule Engine & $S_{code}$ ($\uparrow$), $C_{p}$ ($\uparrow$) \\
 & Ground Truth & $Acc_{seq}$ ($\uparrow$), $Acc_{param}$ ($\uparrow$) \\ 
\cmidrule{1-3}
\textbf{Efficiency} 
 & System Logs & $Succ.$ ($\uparrow$), Tokens ($\downarrow$), Loop Rate ($\downarrow$) \\ 
\bottomrule
\end{tabular}
}
\end{table}

\paragraph{Evaluation Framework}

We evaluate performance across three dimensions:
(1) \textbf{Scientific Validity}: Measured by a composite \textit{Semantic Score} $S_{sem}$ for keyword coverage and an LLM-as-a-Judge \textit{Scientific Score} $C_{s}$ for procedural logic.
(2) \textbf{Physical Compliance}: Quantified by the \textit{Overall Code Score} $S_{code}$, which aggregates \textit{Physical Compliance} $C_{p}$ verified by the symbolic rule engine, along with \textit{Sequence} $Acc_{seq}$ and \textit{Parameter Accuracy} $Acc_{param}$. Notably, $S_{code}$ acts as a hard gatekeeper: schema format failures yield a zero score.
(3) \textbf{System Efficiency}: Tracks \textit{Success Rate} $Succ.$, token consumption \textit{Tokens}, and \textit{Loop Rate}.
Table~\ref{tab:evaluation_metrics} summarizes these metrics; mathematical derivations are detailed in Appendix~\ref{app:metrics_wjy}.


\begin{table*}[t]
\centering
\small
\caption{Main Results on Protocol Drafting (Subset A) and Code Generation (Subset B). \textcolor{green}{$\dagger$}: indicates statistical significance compared to all baselines (Biomni, ReAct, AutoGPT, Reflexion) with $p < 0.001$ (paired t-test).}
\label{tab:main_results}
\vspace{-10pt}
\resizebox{\linewidth}{!}{
\begin{tabular}{@{}llcccc|ccc@{}}
\toprule
 & & \multicolumn{4}{c}{\textbf{Subset A: Scientific Reasoning}} & \multicolumn{3}{c}{\textbf{Subset B: Hardware Execution}} \\
\cmidrule(lr){3-6} \cmidrule(lr){7-9}
\textbf{Method} & \textbf{Backbone} & ROUGE-L $\uparrow$ & $S_{sem} \uparrow$ & \textbf{$C_s\textcolor{green}{\dagger} \uparrow$} & Time (s) $\downarrow$ & $S_{code} \uparrow$ & \textbf{$C_p \uparrow$} & $Acc_{param} \uparrow$ \\ \midrule
Direct & GPT-4o & 0.107 & 0.202 & 0.189 & 13.8 & 0.590 & 0.995 & 0.295 \\
Direct & Gemini-3-Flash & 0.130 & 0.247 & 0.322 & 12.1 & 0.576 & 0.996 & 0.287 \\
Direct & DeepSeek-V3 & 0.123 & 0.260 & 0.285 & 52.1 & 0.495 & 0.995 & 0.205 \\
Direct & Qwen3-30B-a3b & 0.124 & 0.249 & 0.201 & 20.0 & 0.509 & 0.988 & 0.222 \\
Direct & Kimi-k2-Instruct & 0.097 & 0.271 & 0.286 & 83.1 & 0.499 & 0.971 & 0.215 \\ \midrule
Biomni & (Specialized) & 0.081 & 0.252 & 0.342 & 87.1 & N/A & N/A & N/A \\
ReAct & Gemini-3-Flash & 0.116 & 0.268 & 0.455 & \textbf{44.5} & 0.038 & 0.210 & 0.103 \\
Reflexion & Gemini-3-Flash & 0.118 & 0.282 & 0.439 & 148.4 & 0.278 & 0.534 & 0.403 \\
AutoGPT & Gemini-3-Flash & 0.116 & 0.258 & 0.429 & 119.6 & 0.540 & 0.911 & 0.468 \\ \midrule
\rowcolor[HTML]{F2F2F2} 
\textbf{BioProAgent} & Gemini-3-Flash & \textbf{0.147} & \textbf{0.344} & \textbf{0.591} & 71.8 & \textbf{0.653} & \textbf{0.956} & \textbf{0.610} \\ \bottomrule
\end{tabular}
}
\end{table*}

\begin{table*}[t]
\centering
\small
\caption{Main Results on Long-Horizon Stability (Subset C) and Error Correction (Subset D). 
Bold indicates the best performance within agent-based frameworks.}
\label{tab:stability_healing}
\vspace{-10pt}
\resizebox{\linewidth}{!}{
\begin{tabular}{@{}ll ccc | ccc@{}}
\toprule
& & \multicolumn{3}{c}{\textbf{Subset C: Long-Horizon Stability}} & \multicolumn{3}{c}{\textbf{Subset D: Error Correction}} \\
\cmidrule(lr){3-5} \cmidrule(lr){6-8}
\textbf{Method} & \textbf{Backbone} & {Succ.} $\uparrow$ & $Acc_{param}$ $\uparrow$ & \textbf{$C_p$} $\uparrow$ & Correction Succ. $\uparrow$ & \textbf{$C_p$} $\uparrow$ & Loop Rate $\downarrow$ \\ \midrule
ReAct & Gemini-3-Flash & 88.9\% & 0.114 & 0.217 & 0.0\% & 0.000 & 40.0\% \\
Reflexion & Gemini-3-Flash & 33.3\% & 0.000 & 0.000 & 0.0\% & 0.000 & 0.0\% \\
AutoGPT & Gemini-3-Flash & 66.7\% & 0.409 & 0.644 & 0.0\% & 0.000 & 0.0\% \\ \midrule
\rowcolor[HTML]{F2F2F2}
\textbf{BioProAgent} & Gemini-3-Flash & \textbf{100.0\%} & \textbf{0.718} & \textbf{0.950} & \textbf{90.0\%} & \textbf{0.925} & \textbf{0.0\%} \\ \bottomrule
\end{tabular}
}
\end{table*}


\subsection{Main Results}
\subsubsection{Scientific Reasoning (Subset A).}
As shown in Table \ref{tab:main_results}, BioProAgent achieves a Scientific Validity Score ($C_s$) of 0.591, a \textbf{30\% improvement} over the strongest agent baseline (ReAct) and almost double that of vanilla models (Gemini-3-Flash). Notably, even the domain-specialized {Biomni} only reaches a $C_s$ of 0.342, highlighting that domain knowledge alone is insufficient without structured reasoning constraints.
Paired t-tests confirm that these improvements are statistically significant ($p < 0.001$).
BioProAgent's advantage stems from its \textit{Scientific Reflector} within the FSM, which requires a formal review of scientific logic before proceeding to the coding phase, whereas standard agents often neglect crucial controls in pursuit of rapid generation. Cross-model validation (Pearson $r \ge 0.84$) confirms the robustness of our LLM-as-a-Judge metric (see Appendix~\ref{app:significance}).

\subsubsection{Physical Compliance (Subset B).}
Table \ref{tab:main_results} reveals significant differences in code generation performance between the baselines.
{Vanilla LLMs} maintain $C_p = 0.995$ primarily by defaulting to conservative, low-complexity code, which naturally results in poor {Parameter Accuracy} (0.295).
{Standard Agents} (e.g., ReAct), conversely, attempt complex operations but suffer from catastrophic safety failures ($C_p = 0.210$).
\textit{Although ReAct was equipped with the exact same Rule Engine tool, it often fails to invoke these tools proactively due to context saturation or a lack of stopping conditions.}
BioProAgent effectively eliminates this trade-off, achieving the highest $S_{code}$ and $Acc_{param}$ while maintaining high-quality security ($C_p = 0.956$). This confirms FSM's deterministic gating is essential for enforcing safety checks that probabilistic agents tend to bypass.

\begin{figure*}[t]
    \centering
    \begin{subfigure}[b]{0.32\textwidth}
        \centering
        \includegraphics[width=\linewidth]{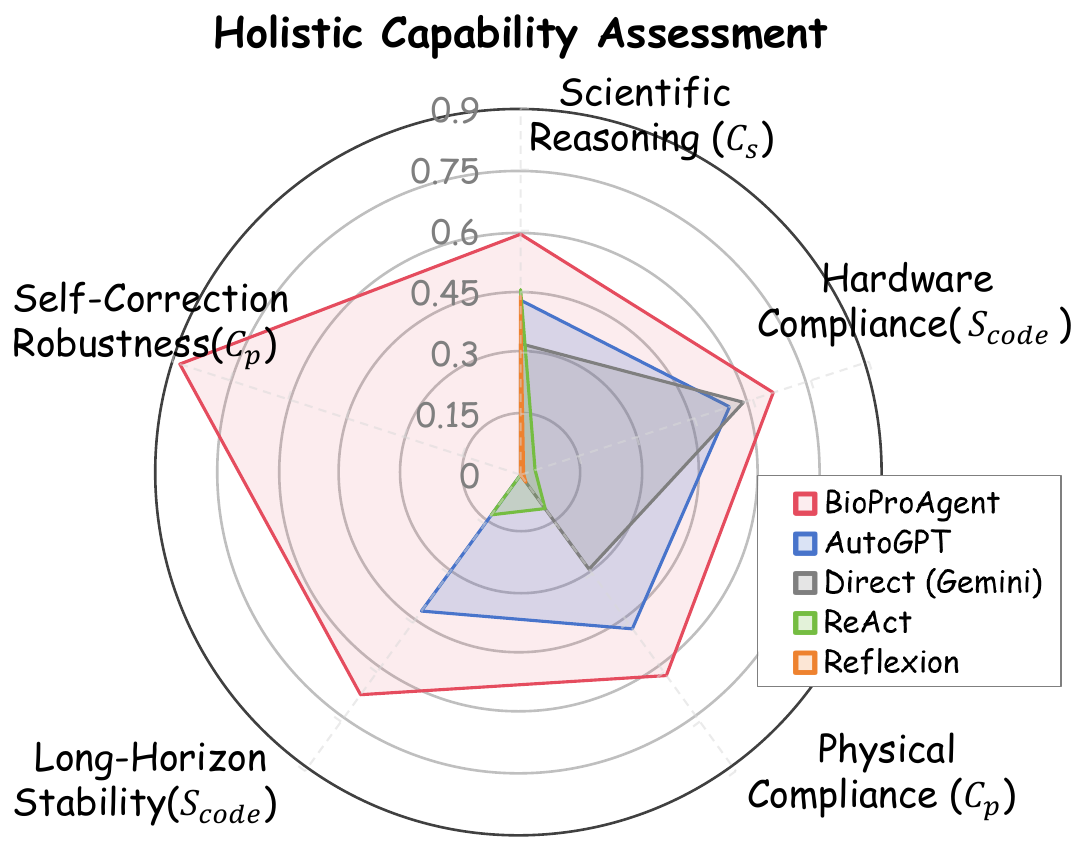}
        \vspace{-15pt}
        \caption{}
        \label{fig:main_results_radar}
    \end{subfigure}
    \hfill 
    \begin{subfigure}[b]{0.32\textwidth}
        \centering
        \includegraphics[width=\linewidth]{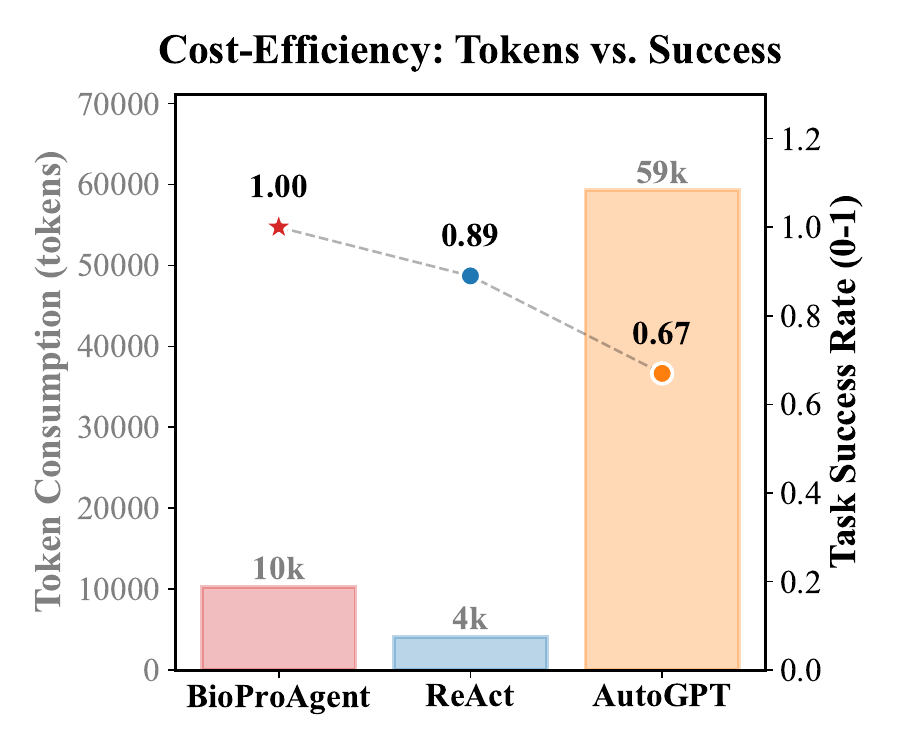}
        \vspace{-15pt}
        \caption{}
        \label{fig:main_results_token}
    \end{subfigure}
    \hfill 
    \begin{subfigure}[b]{0.32\textwidth}
        \centering
        \includegraphics[width=\linewidth]{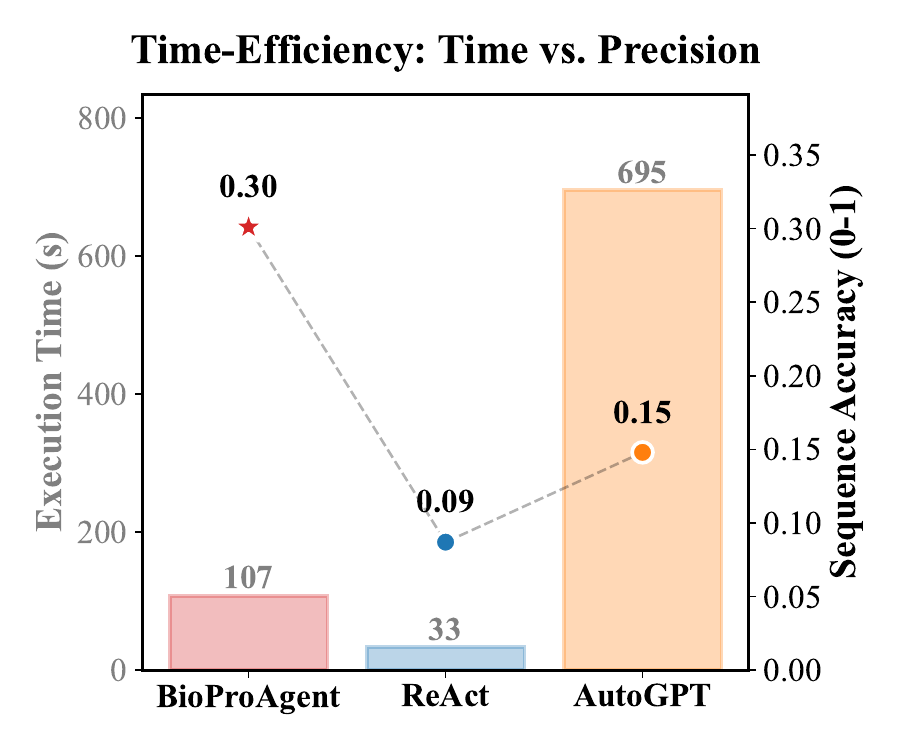}
        \vspace{-15pt}
        \caption{}
        \label{fig:main_results_time}
    \end{subfigure}
    \vspace{-5pt}
    \caption{\textbf{Efficiency and Reliability Analysis.} 
(a) \textbf{Multidimensional Capabilities:} BioProAgent (red) achieves the most expansive capability envelope across five critical dimensions. Note that identical metrics on different axes represent performance in distinct evaluation subsets. 
(b) \textbf{Cost-Efficiency:} On long-horizon tasks (Subset C), our system reduces token consumption by {82\%} compared to AutoGPT while maintaining a 100\% success rate. 
(c) \textbf{Time-Efficiency:} BioProAgent demonstrates superior precision with concise execution time.}
    \label{fig:main_results_combined}
\end{figure*}

\subsubsection{Long-Horizon Stability (Subset C).} 
As shown in Table \ref{tab:stability_healing}, standard agents exhibit significant performance gaps when the workflow expands to an average of 60 steps.
While AutoGPT maintains a 66.7\% success rate, its lack of a structural stopping condition leads to inefficient thought loops and excessive token consumption. 
ReAct achieves an 88.9\% task completion rate, but its low physical consistency and $Acc_{param}=0.114$ reveal ID drift.
\textit{Notably, Reflexion fails completely, resulting in $Acc_{param}=0.000$. This is attributed to ``Schema Corruption'': its verbal feedback mechanism frequently injects conversational artifacts (e.g., apologies) into the JSON payload, causing the strict rules engine to fail to parse the output.}
In contrast, by replacing the high-dimensional data payloads with symbolic pointers, our \textit{Semantic Symbol Grounding} ensures 100\% resource consistency and lossless state tracking.

\subsubsection{Error Correction (Subset D).}
Subset D evaluates the system's ability to diagnose and correct injected physical violations. We define a successful correction as resolving all HALT-level safety threats ($C_p > 0.8$). As shown in Table \ref{tab:stability_healing}, all standard baseline agents exhibit a 0\% correction rate, with ReAct entering infinite retry loops in 40\% of scenarios. To determine if this is merely a tool-use policy failure, we introduced an \textbf{Oracle-ReAct} variant with a forced interlock, prohibiting termination until a ``PASS'' verification signal was received. Oracle-ReAct still achieved a 0\% success rate, falling into a 100\% infinite loop. Without Semantic Symbol Grounding, attempting to feed massive, broken JSON payloads back into the reasoning window caused severe ``Context Paralysis.'' In contrast, BioProAgent successfully restores physical safety ($C_p$) to \textbf{0.925} and achieves an outstanding \textbf{90.0\% }correction success rate. The neuro-symbolic FSM acts as a deterministic router: when a violation is detected by the rules engine, it overwrites the LLM's trajectory and forces a transition to the \texttt{RECTIFY\_CODE} state, aligning outputs with hardware limitations. Appendix~\ref{sec:appendix_error_analysis} provides a detailed analysis of logical errors, resource errors, and physical errors. Furthermore, detailed analysis evaluating whether scaling up to massive ``Thinking'' models can inherently solve this execution gap is provided in Appendix~\ref{sec:app_thinking_models}, further proving the necessity of our framework.


\subsection{Holistic Capability Assessment}
Fig.~\ref{fig:main_results_radar} visualizes the capability envelope of each agent. Crucially, this chart aggregates performance across diverse benchmarks: while $C_p$ and $S_{code}$ appear on multiple axes, they correspond to distinct stress tests, standard generation (Subset B), long-horizon stability (Subset C), and error correction (Subset D). Existing methods exhibit significant capability biases: {Direct Prompting} (Grey) defaults to conservative safety ($C_p$) but fails in complex coordination, while {ReAct} (Green) improves scientific reasoning ($C_s$) at the cost of severe physical hallucinations.
BioProAgent effectively eliminates this trade-off, achieving a balanced pentagonal coverage. The performance gap is most significant in {Long-Horizon} and {Self-Correction} dimensions. Baselines degrade to near-zero performance due to context collapse or open-loop error propagation, BioProAgent maintains robustness, demonstrating that neuro-symbolic FSM is essential for sustaining trustworthy autonomy in high-stakes environments.

\subsection{Efficiency and Cost Analysis}
As shown in Fig.~\ref{fig:main_results_combined}(b), BioProAgent consumes \textbf{82\% fewer tokens} than the AutoGPT on Subset C. This efficiency stems from our \textit{Semantic Symbol Grounding} mechanism, which maintains a near-constant context size by substituting dense data payloads with symbolic pointers, thereby mitigating the "context explosion" inherent in processing dense JSON payloads . Furthermore, Fig.~\ref{fig:main_results_combined}(c) highlights the critical trade-off between execution time and precision. While standard baselines are faster, they lack the requisite precision for complex hardware operations. While the AutoGPT attains moderate precision, its execution time is prohibitively long (avg. 695s). BioProAgent achieves an optimal balance, delivering high-precision results with competitive execution speeds and the widest safety margin ($C_p$), validating practical viability for the real-world laboratory deployment.

\begin{figure}[t]
\centering
\resizebox{\linewidth}{!}{%
\begin{tikzpicture}[
    node distance=1.2cm and 0.8cm, 
    state/.style={rectangle, rounded corners, draw=blue!50!black, fill=blue!5, thick, minimum size=5mm, font=\sffamily\scriptsize, align=center},
    err_state/.style={rectangle, rounded corners, draw=red!60!black, fill=red!5, thick, minimum size=5mm, font=\sffamily\scriptsize, align=center},
    final_state/.style={rectangle, rounded corners, draw=green!40!black, fill=green!5, thick, minimum size=5mm, font=\sffamily\scriptsize, align=center},
    arrow/.style={->, >=stealth, thick, color=gray!70},
    lbl/.style={font=\tiny\sffamily, align=center, color=black!80} 
]

\node[anchor=west, font=\sffamily\scriptsize\bfseries] at (-0.8, 0.8) {(a) Physical Violation ($V > V_{max}$)};

\node[state] (gen1) at (0,0) {DESIGN\\CODE};
\node[lbl, below=0.05cm of gen1] {Output $a_t$:\\\textbf{25,000g}};

\node[err_state] (fix1) [right=of gen1] {RECTIFY\\CODE};
\node[lbl, below=0.05cm of fix1] {Interlock\\($\sigma_{phys}^-$)};

\node[state] (gen2) [right=of fix1] {DESIGN\\CODE};
\node[lbl, below=0.05cm of gen2] {Revised $a_t'$:\\\textbf{15,000g}};

\node[final_state] (end1) [right=of gen2] {SUCCESS};
\node[lbl, below=0.05cm of end1] {Pass $\mathcal{K}_p$\\($\sigma_{phys}^+$)};

\draw[arrow] (gen1) -- node[above, font=\tiny, color=red!70!black] {REJECT} (fix1);
\draw[arrow] (fix1) -- node[above, font=\tiny] {Feedback} (gen2);
\draw[arrow] (gen2) -- node[above, font=\tiny, color=green!50!black] {ACCEPT} (end1);

\node[anchor=west, font=\sffamily\scriptsize\bfseries] at (-0.8, -1.6) {(b) Symbol Grounding Error (Undefined ID)};

\node[state] (b_gen1) at (0,-2.6) {DESIGN\\CODE};
\node[lbl, below=0.05cm of b_gen1] {Output $a_t$:\\\textbf{"new\_plate"}};

\node[err_state] (b_fix1) [right=of b_gen1] {RECTIFY\\CODE};
\node[lbl, below=0.05cm of b_fix1] {Grounding Fail\\(Lookup Error)};

\node[state] (b_gen2) [right=of b_fix1] {DESIGN\\CODE};
\node[lbl, below=0.05cm of b_gen2] {Resolved:\\\textbf{"plate\_1"}};

\node[final_state] (b_end1) [right=of b_gen2] {SUCCESS};
\node[lbl, below=0.05cm of b_end1] {Valid $\in \Omega$};

\draw[arrow] (b_gen1) -- node[above, font=\tiny, color=red!70!black] {REJECT} (b_fix1);
\draw[arrow] (b_fix1) -- node[above, font=\tiny] {Reflect} (b_gen2);
\draw[arrow] (b_gen2) -- node[above, font=\tiny, color=green!50!black] {ACCEPT} (b_end1);

\end{tikzpicture}
}
\vspace{-15pt}
\caption{\textbf{FSM-Driven Self-Correction Trajectories.} (a) Physical Correction. (b) Symbol Grounding Repair.}
\label{fig:correction_trace}
\end{figure}

\subsection{Qualitative Analysis}
\paragraph{Mechanism of Correction.} 
Standard LLM agents typically operate in an open-loop manner: when a hallucinated or dangerous parameter is generated , execution is immediate, inevitably leading to failure. In contrast, Fig.~\ref{fig:correction_trace} illustrates the FSM's correction mechanism in action, effectively intercepting speed limit violations. In the physical violation case (Fig.~\ref{fig:correction_trace}a), the deterministic Rule Engine ($\mathcal{R}_{phys}$) preemptively intercepts this violation ($V > V_{max}$), triggering a forced state transition to \texttt{RECTIFY\_CODE}. This forces the Neural Planner to recognize the specific error signal ($\sigma_{phys}=-1$) and regenerate the code within safe limits (15,000g). Similarly, for a logical hallucination case (Fig.~\ref{fig:correction_trace}b), the system detects semantic drift (an undefined resource ID \texttt{new\_plate}). The FSM prevents this error propagation, guiding the agent to remap the action to the grounded symbol \texttt{plate\_1}.


\begin{table}[t]
\centering
\small
\caption{\textbf{Representative Error Corrections in Subset D.} The FSM interceptor enforces the \textbf{Rectify} phase, guiding the agent from violations to valid execution.}
\label{tab:correction_examples}
\vspace{-5pt}
\resizebox{\linewidth}{!}{
\begin{tabular}{@{}lp{4cm}p{4cm}@{}}
\toprule
\textbf{Constraint} & \textbf{Design Violation ($s_{design}$)} & \textbf{FSM Rectification ($s_{rectify}$)} \\ \midrule
\textbf{Physical ($\mathcal{K}_p$)} & \texttt{centrifuge(speed=`25000g')} & \texttt{centrifuge(speed=`15000g')} \\
\textit{Trace} & \textcolor{red}{$\times$ Exceeds rotor limit (15k)} & \textcolor{green!60!black}{$\checkmark$ Clamped to safety max} \\ \midrule
\textbf{Symbol ($\Phi$)} & \texttt{transfer(source=`buffer\_x')} & \texttt{transfer(source=`trough\_1')} \\
\textit{Trace} & \textcolor{red}{$\times$ Grounding Fail (No ID)} & \textcolor{green!60!black}{$\checkmark$ Remapped to registry} \\ \midrule
\textbf{Causal} & \texttt{seal\_plate(); transfer(...)} & \texttt{transfer(...); seal\_plate()} \\
\textit{Trace} & \textcolor{red}{$\times$ Blocked op (Sealed)} & \textcolor{green!60!black}{$\checkmark$ Reordered dependency} \\ \bottomrule
\end{tabular}
}
\end{table}

\paragraph{Granularity of Rectification.}
Table~\ref{tab:correction_examples} further demonstrates that these corrections are not merely stochastic retries but involve semantic-level reasoning. We observe three distinct categories of rectification within the DVR loop:
(1) \textbf{Safety Alignment:} Adjusting physics parameters to satisfy the hardware constraints defined in the registry ($\Omega$);
(2) \textbf{Resource Grounding:} Resolving hallucinated variables to physically validated slots via the projection function $\Phi$; and
(3) \textbf{Causal Rectification:} Correcting logical dependencies, such as the critical reordering of \texttt{seal\_plate()} operations.
This confirms that BioProAgent's neuro-symbolic core effectively operationalizes a ``Design-Verify-Rectify'' workflow, ensuring trustworthy execution in high-stakes environments.

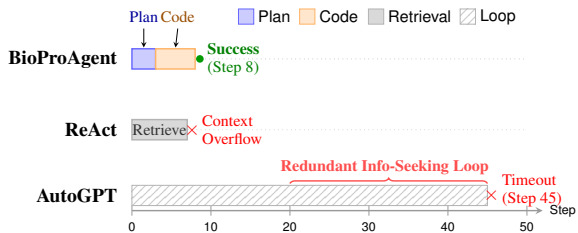
\begin{figure}[t]
\centering
\resizebox{\linewidth}{!}{%
\begin{tikzpicture}[
    x=0.155cm, y=1.0cm, 
    font=\sffamily\scriptsize,
    >=stealth
]

\draw[->, thick, color=gray!80] (0,0) -- (52,0) node[right, color=black] {Step};
\foreach \x in {0,10,20,30,40,50} 
    \draw[gray!80] (\x,0.1) -- (\x,-0.1) node[below, color=black] {\x};

\node[anchor=east, font=\bfseries] at (-1, 3) {\textbf{BioProAgent}};
\draw[dotted, gray!50] (0,3) -- (50,3);

\fill[blue!20, draw=blue!60, thick] (0, 2.8) rectangle (3, 3.2);
\fill[orange!20, draw=orange!60, thick] (3, 2.8) rectangle (8, 3.2);

\node[inner sep=0pt] (plan_center) at (1.5, 3.2) {};
\node[inner sep=0pt] (code_center) at (5.5, 3.2) {};
\draw[<-] (plan_center) -- +(0, 0.4) node[above, font=\small, blue!60!black] {Plan};
\draw[<-] (code_center) -- +(0.3, 0.4) node[above, font=\small, orange!60!black] {Code};

\node[circle, fill=green!60!black, inner sep=1.5pt] (success) at (8.6, 3) {};
\node[green!50!black, right, align=left, font=\small] at (8.7, 3) {\textbf{Success}\\(Step 8)};

\node[anchor=east, font=\bfseries] at (-1, 1.6) {ReAct};
\draw[dotted, gray!40] (0,1.6) -- (50,1.6);

\fill[gray!30, draw=gray!60, thick] (0, 1.4) rectangle (7, 1.8);
\node[font=\small, gray!40!black] at (3.5, 1.6) {Retrieve};

\node[text=red, font=\bfseries] (crash) at (7.6, 1.6) {$\times$};
\node[red, right, align=left, font=\small] at (8, 1.6) {Context\\Overflow};

\node[anchor=east, font=\bfseries] at (-1, 0.3) {AutoGPT};

\fill[pattern=north east lines, pattern color=gray!40, draw=gray!60] (0, 0.1) rectangle (45, 0.5);

\draw[decorate, decoration={brace, amplitude=3pt, raise=1pt}, thick, color=red!70] (20, 0.5) -- (45, 0.5);
\node[above=4pt, color=red!70, align=center, font=\small] at (32, 0.45) {\textbf{Redundant Info-Seeking Loop}};

\node[text=red, font=\bfseries] at (45.5, 0.3) {$\times$};
\node[red, right, align=left, font=\small] at (46, 0.4) {Timeout\\(Step 45)};

\node[anchor=north east] at (50, 4.2) {
    \begin{tikzpicture}[y=0.3cm, x=0.3cm, font=\small]
        \node[anchor=west] at (-2, 0.5) {};
        \fill[blue!20, draw=blue!60] (0,0) rectangle (1,1) node[right, black] at (1,0.5) {Plan};
        \fill[orange!20, draw=orange!60] (4,0) rectangle (5,1) node[right, black] at (5,0.5) {Code};
        \fill[gray!30, draw=gray!60] (8.5,0) rectangle (9.5,1) node[right, black] at (9.5,0.5) {Retrieval};
        \fill[pattern=north east lines, pattern color=gray!40, draw=gray!60] (14.5,0) rectangle (15.5,1) node[right, black] at (15.5,0.5) {Loop};
    \end{tikzpicture}
};

\end{tikzpicture}
}
\vspace{-10pt}
\caption{\textbf{Trace Analysis (Subset C).} Execution divergence on a long-horizon task. TOP: BioProAgent efficiently transitions to coding via FSM. MIDDLE: ReAct fails early due to context overflow. BOTTOM: AutoGPT wastes resources in a retrieval loop.}
\label{fig:trace_divergence}
\end{figure}

\paragraph{Log-Level Behavioral Analysis.}
To diagnose why baseline agents fail in long-horizon tasks despite having access to the same tools, we visualize the execution trajectories of a representative metabolomics workflow in Fig.~\ref{fig:trace_divergence}.
ReAct terminates prematurely (Step 7) as retrieved API schemas saturate its context window, blocking valid action generation.
Lacking a deterministic \textit{stop} signal, AutoGPT enters an infinite loop of redundant \texttt{retrieve\_knowledge} calls (Steps 20--45), consuming excessive tokens without advancing state.
In contrast, BioProAgent's FSM enforces a rigid \textit{Plan $\to$ Code} transition, acting as a cognitive clock that ensures efficient completion in just 8 steps.

\subsection{Ablation Study}
Table~\ref{tab:ablation} quantitatively validates the architectural necessity of each component.
\textbf{Impact of FSM:} Removing the FSM controller leads to a structural failure. The Scientific Validity ($C_s$) degrades to 0.262, while performance on long-horizon tasks (Subset C) collapses to 0.000. Crucially, its physical safety on error-correction tasks (Subset D) drops to 0.782. It is important to note that 0.782 is not a partial success, but rather the baseline physical compliance of the \textit{corrupted inputs themselves}. Because the agent lacks the deterministic state routing (i.e., \texttt{RECTIFY\_CODE}) provided by the FSM, it fails to execute any meaningful self-correction, blindly translating the injected hazards into executable code.
\textbf{Impact of Verification:} Ablating the verifications ($\mathcal{K}_p, \mathcal{K}_s$) causes $C_p$ to drop from 0.956 to 0.902. While less dramatic than the FSM ablation, this deficit demonstrates that neural inference alone, even with advanced prompting, cannot guarantee the zero-defect safety required for irreversible wet-lab automation.
\textbf{Impact of Knowledge \& Disambiguation:} Omitting the disambiguation tool (\texttt{clarify\_experiment\_scope}) or external knowledge retrieval (\texttt{retrieve\_knowledge}) significantly impairs reasoning capabilities. Notably, the sharp decline in $C_s$ ($0.591 \to 0.403$) without clarification underscores the critical role of human-in-the-loop disambiguation for handling under-specified protocols.
Statistical significance is further confirmed via cross-judge verification (Appendix~\ref{app:cross_judge}) and paired t-tests (Appendix~\ref{app:significance}).

\begin{table}[t]
\centering
\caption{Ablation Study of BioProAgent Components.}
\label{tab:ablation}
\vspace{-5pt}
\resizebox{\linewidth}{!}{
\begin{tabular}{@{}lccccc@{}}
\toprule
 & \textbf{Subset A} & \textbf{Subset B } & \textbf{Subset C} & \textbf{Subset D}  \\ 
\textbf{Variant} & $C_s$ & $C_p$ & $S_{code}$ & $C_p$ \\ \midrule
\rowcolor[HTML]{F2F2F2}
\textbf{Full Model} & \textbf{0.591} & \textbf{0.956} & \textbf{0.668} & \textbf{0.925} \\ \midrule
w/o FSM & 0.262 & 0.049 & 0.000 & 0.782 \\
w/o Verification & 0.563 & 0.902 & 0.648 & 0.782 \\
w/o Knowledge & 0.561 & 0.889 & 0.657 & N/A \\ 
w/o Clarify & 0.403 & N/A & 0.632 & N/A \\ \bottomrule
\end{tabular}
}
\end{table}

\section{Conclusion}
We present BioProAgent, a neuro-symbolic framework addressing the critical execution gap in autonomous scientific discovery. By grounding probabilistic LLM reasoning within a deterministic Finite State Machine, BioProAgent enforces a ``Design-Verify-Rectify'' workflow for irreversible wet-lab environments. Evaluations on extended BioProBench demonstrate that BioProAgent achieves SOTA performance and demonstrates robust autonomous self-correction where traditional agents fail. Our work emphasizes that for high-risk physical applications, neural intelligence must be constrained by symbolic safety interlocks to ensure trustworthy and reliable autonomy.


\section*{Limitations \& Future Work}
Despite its performance, BioProAgent has some limitations. First, the framework's security relies on a predefined hardware registry ($\Omega$). While effective for known instruments, this requires manual registration for custom hardware. Future work will explore using a multimodal Large Language Model (MLLM) to automate this process by directly reading the device manual. Additionally, our current evaluation was conducted in high-fidelity simulations using standard APIs. While this validates the correctness of the logic and parameters, real-world physical randomness remains unmodeled. Integrating a real-time visual feedback loop is crucial for managing these physical variables.

\section*{Acknowledgement}
Supported by the China Postdoctoral Science Foundation under Grant Numbers BX20240013 and 2024M760113, the Shenzhen Science and Technology Innovation Commission under Grant KQTD 20240729102051063, and the National Natural Science Foundation of China under Grant Numbers U25B6003 and 62425101.





\bibliography{custom}
\clearpage

\newpage

\appendix

\section*{Appendix}
\label{sec:appendix}

\begin{table*}[b]
    \centering
    \small
    \caption{\textbf{Summary of Notations and Definitions.}}
    \label{tab:notations}
    \renewcommand{\arraystretch}{1.2} 
    \resizebox{\textwidth}{!}{%
    \begin{tabular}{l l p{0.6\textwidth}}
        \toprule
        \textbf{Symbol} & \textbf{Definition} & \textbf{Description} \\
        \midrule
        \multicolumn{3}{l}{\textit{\textbf{Problem Formulation}}} \\
        $I$ & Natural Language Intent & The initial user query describing the experimental objective. \\
        $\mathcal{E}$ & Environmental Context & The set of available lab hardware, reagents, and physical conditions. \\
        $\mathcal{P}$ & Protocol Space & The universal set of all possible protocol sequences. \\
        $P^*$ & Optimal Protocol & The target executable sequence that maximizes scientific validity and physical compliance. \\
        $\mathbb{P}_{\theta}$ & Generation Probability & The probability distribution of generating the complete protocol sequence $P$. \\
        $\mathcal{K}_s$ & Scientific Verification & Function checking experimental logic constraints (enforced by the \textbf{Scientific Reflector}). \\
        $\mathcal{K}_p$ & Physical Verification & Function checking hardware constraints (enforced by the \textbf{Rule Engine} $\mathcal{R}_{phys}$). \\
        $\mathbb{I}[\cdot]$ & Indicator Function & Returns 1 if the logical condition inside holds, 0 otherwise. \\
        
        \midrule
        \multicolumn{3}{l}{\textit{\textbf{Cognitive Memory \& Grounding}}} \\
        $\mathcal{M}$ & Cognitive Memory & The hierarchical memory system $\langle\mathcal{M}_{work}, \mathcal{M}_{episodic}, \mathcal{M}_{long}\rangle$. \\
        $\Phi$ & Projection Function & Maps high-dimensional artifacts to symbolic references with semantic previews. \\
        $\Psi$ & Resolution Function & Dynamically grounds symbolic pointers back to physical parameters ($a_t^{exec} \leftarrow \Psi(a_t)$). \\
        $C_{context}$ & Compressed Context & The lightweight context representation after applying $\Phi$. \\
        $\tau_t$ & Execution Trajectory & The history of FSM-guided states, executed actions, and signals up to time $t$: $[(s_0, a_0, \sigma_0), \dots]$. \\
        
        \midrule
        \multicolumn{3}{l}{\textit{\textbf{Neuro-Symbolic Core (FSM)}}} \\
        $\mathcal{S}$ & State Space & The set of discrete cognitive states (e.g., \textsc{Design}, \textsc{Verify}, \textsc{Rectify}). \\
        $s_t$ & Current State & The active state at time $t$, determining the valid toolset $\mathcal{A}_{valid}$. \\
        $\Sigma$ & Context Signal Space & The mixed boolean/ternary signal space defined as $\{0,1\}^3 \times \{-1,0,1\}^2$. \\
        $\sigma_t$ & Context Signal Vector & A vector $[\sigma_{know}, \sigma_{draft}, \sigma_{code}, \sigma_{sci}, \sigma_{phys}] \in \Sigma$ driving state transitions. \\
        $\Delta$ & Transition Function & The deterministic mapping from current signals and states to the next DVR phase: $\Delta: \Sigma \times \mathcal{S} \rightarrow \mathcal{S}$. \\
        $\pi_{\theta}$ & Neural Policy & The LLM-based single-step planner policy distribution over actions. \\
        $\mathcal{A}$ & Action Space & The complete set of atomic tools available to the agent. \\
        $\Omega$ & Hardware Registry & Structured database of device limits (e.g., max RPM, temperature) defining $\mathcal{K}_p$. \\
        \bottomrule
    \end{tabular}
    }
\end{table*}

\section{Table of Notations}
\label{sec:appendix_notations}
Table~\ref{tab:notations} provides a comprehensive reference for the mathematical notations and symbols employed throughout this paper. To facilitate readability, the symbols are categorized by their functional roles within the BioProAgent framework: \textbf{Problem Formulation}, \textbf{Cognitive Memory}, and the \textbf{Neuro-Symbolic Core}.

\begin{table*}[t!]
\centering
\small
\caption{\textbf{The complete toolset definition.} Tools are clustered by the function in the \textbf{Design-Verify-Rectify (DVR)} workflow. Note that verification tools explicitly implement the $\mathcal{K}_s$ and $\mathcal{K}_p$ functions to drive FSM state transitions.}
\vspace{-5pt}
\label{tab:tool_definitions}
\renewcommand{\arraystretch}{1.1}
\begin{tabularx}{\textwidth}{l p{3.2cm} X}
\toprule
\textbf{Tool Name} & \textbf{Parameters} & \textbf{Function Description} \\
\midrule
\multicolumn{3}{l}{\textit{\textbf{Phase I: Design (Generation \& Alignment)}}} \\
\texttt{generate\_scientific\_draft} & \texttt{query}, \texttt{exp\_info}, \texttt{knowledge} & Generates a natural language experimental protocol draft based on user intent. \\
\texttt{align\_draft\_to\_automation} & \texttt{draft}, \texttt{exp\_info} & Maps the approved draft to specific hardware operations (Intermediate step). \\
\texttt{generate\_machine\_code} & \texttt{aligned\_protocol}, \texttt{suggestion} & Synthesizes executable JSON machine code. Accepts suggestions from the \texttt{Rectify} phase. \\

\midrule
\multicolumn{3}{l}{\textit{\textbf{Phase II: Verify (Hierarchical Validation)}}} \\
\texttt{reflect\_on\_protocol} & \texttt{protocol\_text}, \texttt{query} & (\textbf{Scientific Reflector $\mathcal{K}_s$}) Reviews the draft for logic/controls via CoT. Output updates the ternary signal $\sigma_{sci} \in \{-1, 1\}$. \\
\texttt{validate\_machine\_code} & \texttt{exp\_flow\_json} & (\textbf{Physical Rule Engine $\mathcal{R}_{phys}$}) Evaluates $\mathcal{K}_p$ by checking JSON against the registry $\Omega$. Output updates the ternary signal $\sigma_{phys} \in \{-1, 1\}$. \\

\midrule
\multicolumn{3}{l}{\textit{\textbf{Phase III: Rectify (Self-Correction)}}} \\
\texttt{modify\_protocol} & \texttt{protocol}, \texttt{request} & (\textbf{Logic Rectification}) Revises the draft based on $\mathcal{K}_s$ feedback (triggered when $\sigma_{sci} = -1$). \\
\texttt{fix\_machine\_code} & \texttt{machine\_code}, \texttt{errors} & (\textbf{Physical Rectification}) Autonomously repairs parameters based on error logs from $\mathcal{R}_{phys}$ (when $\sigma_{phys} = -1$). \\

\midrule 
\multicolumn{3}{l}{\textit{\textbf{Auxiliary \& Context}}} \\ 
\texttt{clarify\_experiment\_scope} & \texttt{query}, \texttt{doc\_content} & Resolves ambiguities in initial request before plan begins. \\ 
\texttt{retrieve\_knowledge} & \texttt{query}, \texttt{keywords} & Fetches RAG-based literature/protocols to populate context. \\ 
\texttt{ask\_user\_confirmation} & \texttt{question} & Requests explicit approval before critical execution steps. \\ 
\bottomrule 
\end{tabularx} 
\end{table*}
\vspace{-10pt}
\begin{table*}[t!]
\centering
\small
\renewcommand{\arraystretch}{1.1}
\caption{\textbf{The Priority Decision Matrix defining the FSM transition logic.} $(\sigma_{phys} = -1)$ denotes a physical validation failure, while $(\sigma_{sci} = 1)$ denotes a scientific validation success. Note how the \textbf{[Rectify]} interlocks (Rows 1-2) take strict precedence over \textbf{[Design]} steps.}
\vspace{-5pt}
\label{tab:fsm_logic}
\begin{tabularx}{\textwidth}{c l l X}
\toprule
\textbf{Priority} & \textbf{Signal Condition ($\sigma_t$)} & \textbf{Target State ($s_{t+1}$)} & \textbf{Logic Description (DVR Flow)} \\
\midrule
1 & $(\sigma_{phys} = -1) \land \sigma_{code}$ & \texttt{RECTIFY\_CODE} & \textbf{[Rectify] Safety Interlock}: If Rule Engine ($\mathcal{K}_p$) detects a physical violation (Severity: HALT), immediately force a fix. \\
\hline
2 & $(\sigma_{sci} = -1) \land \sigma_{draft}$ & \texttt{RECTIFY\_DRAFT} & \textbf{[Rectify] Scientific Review}: If Reflector ($\mathcal{K}_s$) rejects the protocol logic (e.g., missing controls), require revision. \\
\hline
3 & $\neg \sigma_{draft} \land \sigma_{know}$ & \texttt{DESIGN\_DRAFT} & \textbf{[Design] Knowledge Ready}: If knowledge is retrieved but no draft exists, enter drafting mode. \\
\hline
4 & $\sigma_{draft} \land (\sigma_{sci} = 1)$ & \texttt{DESIGN\_CODE} & \textbf{[Design] Code Generation}: Synthesize executable machine code directly from the verified and approved scientific draft. \\
\hline
5 & $\sigma_{draft} \land (\sigma_{sci} = 0)$ & \texttt{VERIFY\_DRAFT} & \textbf{[Verify] Draft Review}: Evaluate the newly generated protocol draft against scientific principles (Pending state). \\
\hline
6 & $\sigma_{code} \land (\sigma_{phys} = 1)$ & \texttt{SUCCESS} & \textbf{[Finish] Completion}: Code exists and passes all physical validations ($\mathcal{K}_p$). Task complete. \\
\hline
7 & $\sigma_{ambiguous}$ & \texttt{CLARIFY\_INTENT} & \textbf{[Aux] Ambiguity Check}: If initial intent is unclear, loop back to ask user for parameters. \\
\hline
8 & (Default) & $s_t$ (Self-loop) & \textbf{[Wait]}: If no condition matches, maintain current state (e.g., during retrieval or accumulation). \\
\bottomrule
\end{tabularx}
\end{table*}

\section{Detailed Toolset}
\label{app:tools}
To ensure fairness and reproducibility, all agents in our experiments are equipped with the identical set of operational tools. Table~\ref{tab:tool_definitions} details the specifications, categorized by their functional role in the \textbf{Design-Verify-Rectify (DVR)} workflow.

\section{FSM Decision Logic}
\label{app:fsm}
The Finite State Machine (FSM) acts as the deterministic backbone of BioProAgent, orchestrating the DVR loop. Table~\ref{tab:fsm_logic} details the \textbf{Priority Decision Matrix} used to compute state transitions $s_{t+1} = \Delta(\sigma_t, s_t)$. The rules are evaluated in descending order of priority (1 is highest), ensuring that safety interlocks always override generative actions.

\begin{table*}[t]
\centering
\small
\caption{\textbf{Representative Action Schemas from the Hardware Registry ($\Omega$).} The Physical Rule Engine ($\mathcal{K}_p$) validates generated JSON code against these parameter constraints and physical interlocks.}
\label{tab:action_schema}
\renewcommand{\arraystretch}{1.2} 
\resizebox{\textwidth}{!}{%
\begin{tabular}{l|l|p{6cm}|p{5cm}}
\toprule
\textbf{Device Category} & \textbf{Action Primitive} & \textbf{Parameters (Type: Constraint)} & \textbf{Physical Rules (Examples)} \\ \midrule

\multirow{4}{*}{\textbf{Centrifugation}} & \texttt{Centrifuge} & 
$\bullet$ \texttt{speed\_g} (int: $500 -- 15000$) \newline
$\bullet$ \texttt{time} (duration: HH:MM:SS) \newline
$\bullet$ \texttt{temp\_C} (int: $4 -- 40$) \newline
$\bullet$ \texttt{brake} (enum: off, slow, fast) & 
1. Speed must not exceed rotor limit ($V \le V_{max}$). \newline
2. Temperature requires pre-cooling if $T < 25^{\circ}$C. \\ \hline

\multirow{4}{*}{\textbf{Thermal Control}} & \texttt{Incubate} & 
$\bullet$ \texttt{temp\_C} (int: $4 -- 99$) \newline
$\bullet$ \texttt{shake\_rpm} (int: $0 -- 2000$) \newline
$\bullet$ \texttt{duration} (time) & 
1. Shaking prohibited if plate is unsealed. \newline
2. Max temp depends on plate material (PS vs. PP). \\ \hline

\multirow{4}{*}{\textbf{Liquid Handling}} & \texttt{Transfer} & 
$\bullet$ \texttt{source} (ID string) \newline
$\bullet$ \texttt{dest} (ID string) \newline
$\bullet$ \texttt{volume\_uL} (float: $0.5 -- 1000$) \newline
$\bullet$ \texttt{tip\_type} (enum: p20, p300, p1000) & 
1. Source volume must be $>$ aspiration volume + dead volume. \newline
2. Tip type must match volume range. \\ \hline

\multirow{3}{*}{\textbf{PCR Cycling}} & \texttt{Thermocycle} & 
$\bullet$ \texttt{lid\_temp} (int: $95 -- 105$) \newline
$\bullet$ \texttt{stages} (list of [temp, time, cycles]) & 
1. Lid temp must be $>$ reaction temp to prevent condensation. \\ \bottomrule
\end{tabular}%
}
\end{table*}

\section{Hardware Registry and Action Space Schema}
\label{app:hardware}

To support reproducibility and the reconstruction of the \textbf{Physical Rule Engine ($\mathcal{K}_p$)}, we formally define the device action space. The Hardware Registry ($\Omega$) contains 22 specific instruments. Every action $a_t$ generated by the agent must strictly conform to the parameterized schemas defined below.

\subsection{Device Categories and Functions}
The registry organizes hardware into four functional categories to enable end-to-end wet-lab automation:
\begin{itemize}
    \item \textbf{Liquid Handling:} Automated pipetting workstations, acoustic liquid handlers, and dispensers.
    \item \textbf{Thermal Control:} PCR thermal cyclers, incubators, and plate sealers.
    \item \textbf{Separation \& Processing:} Centrifuges, shakers, and magnetic separators.
    \item \textbf{Analysis:} Multi-mode plate readers, qPCR systems, and electrophoresis imagers.
\end{itemize}

\subsection{Action Primitives and Parameter Constraints}
Table \ref{tab:action_schema} specifies the schemas for representative action primitives. All constraints (e.g., integer ranges, enumerations) are strictly enforced by the \textbf{Physical Rule Engine ($\mathcal{K}_p$)} during the Verification phase of the DVR loop.

\subsection{Full Registry Availability}
Due to space constraints, Table \ref{tab:action_schema} presents a subset of the 22 devices. The complete JSON schema definitions, including configurations for the Automated Nitrogen Evaporator, Microbial Colony Picker, Plate Washer, are provided in the codebase.

\section{Detailed Evaluation Metrics}
\label{app:metrics_wjy}
To rigorous assess the quality of generated protocols and executable machine code, we implement a hybrid evaluation framework that integrates semantic similarity analysis, deterministic rule-based verification, and expert-aligned LLM-as-a-Judge assessments.

\subsection{Protocol Generation Metrics}
Let $P_{gt} = \{p_{1}, ..., p_{k}\}$ denote the ground truth protocol sequence, and $P_{gen} = \{p'_{1}, ..., p'_{m}\}$ denote the generated sequence.

\paragraph{Text and Semantic Metrics.}
We employ three complementary metrics to measure lexical coverage and structural alignment:

\noindent{1. Keyword F1 ($F1_K$):} Utilizing KeyBERT~\cite{keybert}, we extract the top-$k$ ($k=64$) domain-specific entities (e.g., reagents, instruments). Precision ($P_K$) and Recall ($R_K$) are computed based on the entity overlap between $P_{gt}$ and $P_{gen}$:
\begin{equation}
    F1_K = \frac{2 \cdot P_K \cdot R_K}{P_K + R_K}.
\end{equation}

\noindent{2. Step Recall ($SR$)~\cite{bioprotocolbench2025}:} To quantify structural completeness, we measure the proportion of ground truth steps that are semantically recovered in the generation:
\begin{equation}
    SR = \frac{\sum_{s \in P_{gt}} \mathbb{I}\left[\max_{s' \in P_{gen}} \text{Sim}(s, s') \geq \tau\right]}{|P_{gt}|},
\end{equation}
where $\text{Sim}(\cdot)$ denotes the cosine similarity of sentence embeddings derived from \texttt{all-mpnet-base-v2}, with a semantic threshold $\tau=0.7$.

\noindent{3. Semantic Consistency Score ($S_{sem}$):} We implement regex-based extraction to isolate experimental verbs ($\mathcal{V}$) and numerical parameters ($\mathcal{PR}$). The component scores are defined as:
\begin{equation}
\resizebox{0.89\hsize}{!}{$
    Cov_{\text{step}} = \frac{|\mathcal{V}_{gt} \cap \mathcal{V}_{gen}|}{|\mathcal{V}_{gt}|}, \quad
    E_{\text{param}} = \frac{|\mathcal{PR}_{gt} \cap \mathcal{PR}_{gen}|}{|\mathcal{PR}_{gt}|}.
    $}
\end{equation}
To ensure numerical stability, if $|\mathcal{V}_{gt}|=0$ or $|\mathcal{PR}_{gt}|=0$, the respective sub-score defaults to 0.5. The aggregated semantic score is computed as:
\begin{equation}
\begin{split}
    S_{sem} &= 0.15 \cdot Cov_{\text{step}} + 0.15 \cdot E_{\text{param}} \\
            &\quad + 0.30 \cdot F1_{K} + 0.40 \cdot SR.
\end{split}
\end{equation}
\textbf{Rationale for Weights:} These coefficients were calibrated based on consultations with senior wet-lab biologists to prioritize structural completeness ($SR$) and keyword accuracy ($F1_K$) over minor syntactic variations. This ensures the metric reflects practical laboratory utility rather than mere linguistic fluency.

\paragraph{Scientific Validity ($C_s$).}
Since semantic overlap may miss subtle logical flaws, we leverage an LLM-as-a-Judge (Gemini-3-Flash) to audit protocols across five weighted dimensions: Critical Step Coverage ($S_{cov}$), Parameter Precision ($S_{param}$), Control Fidelity ($S_{ctrl}$), Biosafety ($S_{safe}$), and Objective Alignment ($S_{obj}$).
\begin{equation}
\begin{split}
    C_{s} &= 0.3 S_{cov} + 0.2 S_{param} + 0.15 S_{ctrl} \\
          &\quad + 0.1 S_{safe} + 0.25 S_{obj}.
\end{split}
\end{equation}

\subsection{Machine Code Generation Metrics}
Let $S_{gt}$ and $S_{gen}$ represent the sequences of executable operations. Code quality is evaluated through a three-layered approach:

\paragraph{Rule-Based Validity.}
We strictly verify basic executability via deterministic checks:

\noindent{1. Schema Compliance ($Schema_{comp}$):} The proportion of mandatory fields (e.g., node types, connection IDs) that adhere to the JSON definition.

\noindent{2. Resource Validity ($Res_{valid}$):} The ratio of valid instrument IDs referenced in the code:
\begin{equation}
    Res_{valid} = \frac{1}{M} \sum_{i=1}^M \mathbb{I}[n_i^{\text{id}} \in \Omega_{valid} \cup \{-1\}],
\end{equation}
where $\Omega_{valid}$ represents the authorized hardware registry and $-1$ denotes manual steps.

\noindent{3. Sequence Accuracy ($Acc_{seq}$):} The Longest Common Subsequence (LCS) ratio measuring the ordering correctness of functional operations:
\begin{equation}
    Acc_{seq} = \frac{2 \cdot |\text{LCS}(S_{gt}, S_{gen})|}{|S_{gt}| + |S_{gen}|}.
\end{equation}

\paragraph{Physical Compliance ($C_p$).}
A symbolic rule engine validates $N$ hardware constraints (e.g., $rpm \le 15,000$). The score penalizes Critical Errors (HALT, $\alpha=0.2$) and Warnings (WARN, $\beta=0.05$):
\begin{equation}
    C_{p} = \max\left(0, 1 - ({\alpha N_{halt} + \beta N_{warn}})\right).
\end{equation}
This penalty mechanism ensures that safety violations significantly degrade the score, aligning with the zero-tolerance policy for hardware damage.

\paragraph{Parameter Accuracy ($Acc_{param}$).}
Fine-grained parameter correctness is evaluated by LLM-as-a-Judge across Node Completeness ($S_{n\text{-}com}$), Value Accuracy ($S_{p\text{-}acc}$), and Field Completeness ($S_{p\text{-}com}$):
\begin{equation}
    Acc_{param} = 0.3 S_{n\text{-}com} + 0.4 S_{p\text{-}acc} + 0.3 S_{p\text{-}com}.
\end{equation}

\paragraph{Overall Code Score ($S_{code}$).}
We employ a multiplicative formulation where $Schema_{comp}$ acts as a soft gating factor. Unlike additive metrics where accurate parameters might mask a broken syntax, this design ensures that if the foundational schema structure is flawed ($Schema_{comp} \rightarrow 0$), the overall executability score $S_{code}$ is penalized towards zero. This reflects the operational reality that syntactically invalid code is non-executable regardless of semantic intent.
\begin{equation}
\begin{split}
    &S_{code} = Schema_{comp} \cdot \Big( 0.35 Acc_{seq} +  \\
    &0.25 Acc_{param} + 0.30 C_{p} + 0.10 Res_{valid} \Big).
\end{split}
\end{equation}

\subsection{System Performance Metrics}
To assess operational viability, we track:
\textbf{Efficiency:} Execution Time ($T$), Total Steps ($N_{step}$), and Token Consumption ($N_{token}$).
\textbf{Success Rate ($Succ.$):} A task is deemed successful only if it meets strict domain validity criteria ($S_{code} \ge \tau_{code}$ and $C_p=1.0$):
\begin{equation}
    Succ. = \frac{\sum_{i=1}^{N_{total}} \mathbb{I}[\text{Task}_i \text{ is Valid}]}{N_{total}}.
\end{equation}

\subsection{Statistical Analysis \& Metric Verification}
To ensure the rigor of our evaluation, we conducted two types of analyses: (1) Cross-model verification to assess the objectivity of our LLM-as-a-Judge metric, and (2) Paired t-tests to determine the statistical significance of our performance gains.

\subsubsection{Metric Reliability (Cross-Judge)}
\label{app:cross_judge}

\begin{table}[h]
\centering
\small
\caption{\textbf{Multi-Judge Consistency Analysis.} Comparison of $C_s$ scores across three distinct LLMs ($N=640$ evaluations). GPT-4o confirms the trends with the highest correlation, validating the robustness of our metric.}
\label{tab:cross_judge_multi}
\resizebox{\linewidth}{!}{
\begin{tabular}{l c c c}
\toprule
\textbf{Model Variant} & \textbf{Gemini} & \textbf{Kimi} & \textbf{GPT-4o} \\
\midrule
BioProAgent-Full & 0.591 & 0.597 & 0.515 \\
w/o Knowledge & 0.561 & 0.552 & 0.497 \\
w/o Clarify & 0.403 & 0.421 & 0.374 \\
w/o FSM & 0.262 & 0.285 & 0.105 \\
\midrule
\textbf{Correlation ($r$) \newline 
w/ Gemini} & \textbf{1.00} & \textbf{0.84} & \textbf{0.91} \\
\bottomrule
\end{tabular}
}
\end{table}

To rule out self-preference bias in the Gemini-based judge ($C_s$), we re-evaluated the ablation dataset using GPT-4o~\cite{gpt-4o} and Kimi-k2-Instruct~\cite{kimi-k2}. As shown in Table~\ref{tab:cross_judge_multi}, the scoring trends across different reasoning engines are highly correlated ($r \approx 0.91$), confirming that the $C_s$ metric objectively reflects protocol quality. GPT-4o exhibits a near-perfect correlation with Gemini ($r=0.91$), validating the reliability of our primary metric. All three judges consistently rank {BioProAgent-Full} as the top performer and identify the removal of the FSM (w/o FSM) as the most detrimental ablation. Notably, GPT-4o assigns significantly lower scores to the w/o FSM variant ($0.105$) compared to Gemini ($0.262$), suggesting that advanced models penalize the lack of structured planning even more severely.

\subsubsection{Statistical Significance (P-values)}
\label{app:significance}

\begin{table*}[t!]
\centering
\small
\renewcommand{\arraystretch}{1.15}
\caption{Overview of the \textbf{Extended BioProBench}.}
\resizebox{\textwidth}{!}{
\begin{tabular}{l l l l} 
\toprule
\textbf{Benchmark Subset} & \textbf{Scale \& Granularity} & \textbf{Evaluation Focus} & \textbf{Key Metrics} \\ 
\midrule
\textbf{A: Scientific Drafting} & 160 Protocols & \textit{Intent Reasoning} & \textbullet\ Intent Alignment \\
(Cross-domain) & (approx. 3k tokens/doc) & Knowledge retrieval & \textbullet\ Scientific Logic Score \\
\midrule
\textbf{B: Automation Conversion} & 41 Sub-Experiments & \textit{Local Code Mapping} & \textbullet\ Parameter Accuracy \\
(Syn-Bio Focus) & (22 Device APIs) & Schema compliance & \textbullet\ API Validity \\
\midrule
\textbf{C: Long-Horizon Exec.} & \textbf{9 Protocols} & \textit{Global Orchestration} & \textbullet\ \textbf{Global Success Rate} \\
(Full Pipeline) & \textbf{(547 Total Steps)} & \textbf{State Persistence (Avg. 60.8 steps)} & \textbullet\ ID Consistency \\
\midrule
\textbf{D: Error Correction} & 30 Injection Cases & \textit{Deterministic Robustness} & \textbullet\ Detection Recall \\
(Robustness) & (Physical/Logic Errors) & Self-healing capabilities & \textbullet\ Fix Success Rate \\
\bottomrule
\end{tabular}
}
\label{tab:dataset_summary}
\end{table*}

\begin{table}[h]
\centering
\caption{\textbf{Statistical Significance Analysis (Paired t-test).} Comparison of BioProAgent against baselines and ablation variants ($N=160$). All improvements are highly statistically significant.}
\label{tab:significance_all}
\resizebox{\linewidth}{!}{
\begin{tabular}{lp{0.36\linewidth}cc}
\toprule
\textbf{Category} & \textbf{Comparison Pair \newline (vs. Full Model)} & \textbf{Mean Diff.} & \textbf{P-value} \\
\midrule
\multirow{3}{*}{\textbf{Baselines}} 
 & ReAct & +0.136 & $< 10^{-10}$ \\
 & AutoGPT & +0.162 & $< 10^{-11}$ \\
 & Reflexion & +0.152 & $< 10^{-11}$ \\
\midrule
\multirow{3}{*}{\textbf{Ablations}} 
 & w/o Knowledge & +0.030 & $0.025^{*}$ \\
 & w/o Clarify & +0.188 & $< 10^{-13}$ \\
 & w/o FSM & +0.329 & $< 10^{-49}$ \\
\bottomrule
\multicolumn{4}{p{\linewidth}}{\footnotesize $^{*}$ Significant ($p < 0.05$); \newline All others are extremely significant ($p < 0.001$).} \\
\end{tabular}
}
\end{table}

We performed paired sample t-tests on the Scientific Validity scores ($C_s$) for the entire Subset A ($N=160$). Table~\ref{tab:significance_all} reports the results for both the main baselines and the ablation variants.

\section{Benchmark Details}

As shown in Table~\ref{tab:dataset_summary}, we summary the extended BioProBench.
\textbf{Subset A (Protocol Drafting)} 160 protocols across 15 biological domains with varying query richness to test intent disambiguation and scientific reasoning;
\textbf{Subset B (Code Generation)} 41 synthetic biology sub-experiments with paired natural language drafts and ground-truth machine codes to evaluate hardware schema compliance and parameter accuracy;
\textbf{Subset C (Long-Horizon)} tests 9 end-to-end long protocols, featuring up to 71 major steps, designed to test the limits of state persistence and context management;
\textbf{Subset D (Error Correction)} measures 30 code snippets with injected physical/logic errors to evaluate the system's deterministic self-healing capability. 
We enriched the environment with API schemas for 22 specialized synthetic biology devices and a comprehensive library of consumable identifiers.

\subsection{Statistical Significance of Subset C} 

\begin{table}[h]
\centering
\small
\caption{\textbf{Complexity Statistics of the 9 Long-Horizon Tasks in Subset C.} Unlike standard benchmarks with short trajectories, these tasks involve up to 238 atomic steps and 169 unique consumables. A failure at any step leads to task failure.}
\label{tab:subset_c_stats}
\renewcommand{\arraystretch}{1.12}
\resizebox{\linewidth}{!}{%
\begin{tabular}{l p{0.25\linewidth} p{0.25\linewidth} p{0.25\linewidth}}
\toprule
\textbf{Task ID} & \textbf{{Device \newline Interaction \newline Nodes}} & \textbf{{Atomic \newline Operation \newline Steps}} & \textbf{{Involved  \newline Consumables}} \\
\midrule
Sample 1 & 15 & 55 & 31 \\
Sample 2 & 4 & 24 & 6 \\
Sample 3 & 8 & 16 & 17 \\
Sample 4 & 17 & 35 & 26 \\
Sample 5 & 18 & 55 & 42 \\
Sample 6 & 19 & 27 & 31 \\
\textbf{Sample 7} & \textbf{71} & \textbf{238} & \textbf{130} \\
Sample 8 & 36 & 81 & 169 \\
Sample 9 & 8 & 16 & 38 \\
\midrule
\textbf{Total} & \textbf{196} & \textbf{547} & \textbf{490} \\
\textbf{Average} & \textbf{21.8} & \textbf{60.8} & \textbf{54.4} \\
\bottomrule
\end{tabular}
}
\end{table}

While Subset C contains 9 high-level tasks, we argue that the sample size should be interpreted through the lens of trajectory complexity rather than task count. As detailed in Table~\ref{tab:subset_c_stats}, these are not single-turn queries but deep sequential decision processes. The tasks comprise a total of 547 atomic operation steps (avg. 60.8, max 238 per task). Executing Sample 7 requires managing 130 distinct consumables and 71 device interactions without a single hallucination. In wet-lab protocols, steps are strictly coupled. A deviation at Step 10 invalidates the result at Step 238.

Therefore, BioProAgent's 100\% success rate in this subset represents 547 consecutive correct decisions in a grounded physical environment. If we assume a baseline agent has even a 95\% step-wise accuracy, the probability of successfully completing the longest task (Sample 7, 238 steps) is negligible ($0.95^{238} \approx 5 \times 10^{-6}$). Our result demonstrates that the FSM-based architecture effectively prevents the compound error explosion typical in probabilistic LLMs.

\begin{table*}[t]
\centering
\small 
\renewcommand{\arraystretch}{1.3} 
\caption{\textbf{Gallery of Autonomous Corrections.} The table compares the raw design violation (left), the specific safety interlock triggered by the Rule Engine (center), and the rectified code after FSM-guided feedback (right). These examples correspond directly to the qualitative traces discussed in Section~4.5.}
\label{tab:correction_gallery}
\begin{tabularx}{\textwidth}{l X p{0.3\textwidth} X}
\toprule
\textbf{Constraint Type} & \textbf{Design Violation ($s_{design}$)} & \textbf{Interlock Trigger ($\neg \mathcal{K}_p$)} & \textbf{FSM Rectification ($s_{rectify}$)} \\ \midrule

\textbf{Safety Alignment} & 
\texttt{op.centrifuge(} \newline
\texttt{\ \ speed=25,000g,} \newline 
\texttt{\ \ duration=15min)} & 
\textcolor{red}{\textbf{[HALT]} Speed 25,000g exceeds rotor limit (Max: 15,000g).} & 
\texttt{op.centrifuge(} \newline
\texttt{\ \ speed=15,000g,} \newline 
\texttt{\ \ duration=15min)} \newline 
\textcolor{green!60!black}{\textit{\# Clamped to safe limit}} \\ \hline

\textbf{Resource Grounding} & 
\texttt{op.transfer(} \newline
\texttt{\ \ source=new\_plate,} \newline 
\texttt{\ \ vol=50)} & 
\textcolor{red}{\textbf{[HALT]} Resource ID \texttt{new\_plate} not found in Registry $\Omega$.} & 
\texttt{op.transfer(} \newline
\texttt{\ \ source=plate\_1,} \newline 
\texttt{\ \ vol=50)} \newline 
\textcolor{green!60!black}{\textit{\# Mapped to valid grounded ID}} \\ \hline

\textbf{Causal Rectification} & 
\texttt{op.seal\_plate()} \newline 
\texttt{op.add\_reagent(vol=10)} & 
\textcolor{red}{\textbf{[HALT]} Cannot \texttt{add\_reagent} to a sealed container (Physical Collision Risk).} & 
\texttt{op.add\_reagent(vol=10)} \newline 
\texttt{op.seal\_plate()} \newline 
\textcolor{green!60!black}{\textit{\# Sequence causally reordered}} \\ \bottomrule

\end{tabularx}
\end{table*}

\subsection{Subset D: Error Injection Taxonomy}
\label{sec:appendix_error_analysis}

To further investigate the robustness of BioProAgent, we present a granular breakdown of performance metrics across three specific error categories in Subset D (Error Correction). Table~\ref{tab:error_breakdown} details the agent's behavior when correcting Resource Grounding, Safety Alignment, and Causal Rectification.

\begin{table}[h]
    \centering
    \caption{Granular Performance Breakdown by Error Type (Subset D). While syntactic perfection remains challenging in complex repairs, BioProAgent maintains high $C_p$, reinforcing its role as a safety interlock.}
    \label{tab:error_breakdown}
    \resizebox{1\linewidth}{!}{%
    \begin{tabular}{l c c c c}
        \toprule
        \textbf{Error Type} & \textbf{Counts} & \textbf{$C_p$} & \textbf{LCS Recall} & \textbf{$Acc_{seq}$} \\
        \midrule
        Resource Grounding & 10 & 0.935 & 1.000  & 0.999 \\
        Safety Alignment & 10 & 0.970 & 1.000  & 0.979 \\
        Causal Rectification & 10 & 0.870 & 0.900 & 0.897 \\
        \bottomrule
    \end{tabular}
    }
\end{table}

As shown in Table~\ref{tab:error_breakdown}, BioProAgent demonstrates exceptionally high \textit{Physical Compliance} ($C_p = 0.935$) and flawless \textit{LCS Recall} (1.000) for resource hallucinations. This indicates that while the agent correctly identifies the semantic intent and enforces safety limits, strictly formatting the JSON schema during complex repairs remains a non-trivial challenge. Crucially, the high $C_p$ scores confirm that the neuro-symbolic FSM successfully prevents dangerous instructions even when code syntax is imperfect.

\subsection{The Error Correction Gallery (Subset D)}
\label{app:error_gallery}

Table~\ref{tab:correction_gallery} details specific instances of \textbf{autonomous rectification} captured during the ablation study (corresponding to the taxonomy in Appendix~\ref{sec:appendix_error_analysis}). These examples illustrate the three primary categories of hard constraints enforced by our \textbf{Physical Rule Engine ($\mathcal{K}_p$)}, demonstrating how the FSM deterministically intercepts hazardous instructions before physical execution.

\section{Impact of Scaling to ``Thinking'' Models}
\label{sec:app_thinking_models}

\begin{table*}[ht]
\centering
\small
\caption{Evaluation of ``Thinking'' Reasoning Models on Subsets C and D.}
\label{tab:app_thinking_models}
\vspace{-5pt}
\resizebox{\linewidth}{!}{
\begin{tabular}{@{}ll cc | ccc@{}}
\toprule
& & \multicolumn{2}{c}{\textbf{Subset C}} & \multicolumn{3}{c}{\textbf{Subset D}} \\
\cmidrule(lr){3-4} \cmidrule(lr){5-7}
\textbf{Method} & \textbf{Backbone} & Succ. $\uparrow$ & $Acc_{param}$ $\uparrow$ & Corr. Succ. $\uparrow$ & $C_p$ $\uparrow$ & Time (s) $\downarrow$ \\ \midrule
Direct & DeepSeek-V3.2-Think & 100.0\% & 0.414 & 93.3\% & 0.948 & 196.2 \\
Direct & Gemini-3-Pro-Think & 100.0\% & 0.439 & 73.3\% & 0.765 & 81.4 \\ \midrule
BioProAgent & DeepSeek-V3.2-Think & 55.6\% & 0.366 & \textbf{93.3\%} & \textbf{0.955} & 1313.3 \\
BioProAgent & Gemini-3-Pro-Think & 44.4\% & 0.256 & 90.0\% & 0.922 & 316.1 \\ \midrule
\rowcolor[HTML]{F2F2F2}
\textbf{BioProAgent} & Gemini-3-Flash & \textbf{100.0\%} & \textbf{0.718} & 90.0\% & 0.925 & \textbf{128.0} \\ \bottomrule
\end{tabular}
}
\end{table*}

A critical question regarding the ``Execution Gap'' is whether scaling up to massive, state-of-the-art ``System 2'' reasoning models (e.g., DeepSeek-V3.2-Think~\cite{deepseekv3.2}, Gemini-3-Pro-Think~\cite{gemini3pro}) can inherently solve the alignment and error correction challenges, rendering neuro-symbolic frameworks unnecessary. To investigate this, we evaluated these models on Subsets C and D, both via direct prompting and as the cognitive engine inside BioProAgent.

\paragraph{The Capability Plateau.} 
As shown in Table \ref{tab:app_thinking_models}, massive reasoning models exhibit strong single-shot robustness due to their extensive internal \texttt{<think>} trajectories. Direct DeepSeek-V3.2-Think achieves a 93.3\% correction rate on its own. However, relying solely on massive models scales poorly for iterative robotic tasks and fails to capture the fine-grained sequence logic in long-horizon tasks (achieving only $Acc_{param}=0.414$ on Subset C).

\paragraph{Reasoning Context Overflow.} 
Counter-intuitively, when integrating these heavy reasoning models inside BioProAgent's iterative loop, performance on long-horizon tasks degraded (Subset C Success Rate dropped to 44\%-55\%). BioProAgent's FSM acts as an external multi-step framework. When models generate thousands of tokens of internal \texttt{<think>} traces at every FSM state transition, the working memory rapidly saturates (averaging $>66,000$ tokens). This causes ``Attention Dilution,'' leading the model to forget strict JSON formatting constraints and fail prematurely.

\paragraph{The Efficiency Triumph.} 
The true value of our neuro-symbolic architecture lies in democratizing high-fidelity autonomy. By providing a deterministic external reasoning track, BioProAgent empowers a highly efficient, lightweight model (Gemini-3-Flash) to achieve a 90.0\% correction rate ($C_p = 0.925$). Crucially, BioProAgent with Flash resolves physical violations in an average of \textbf{128.0 seconds}, whereas BioProAgent with DeepSeek-V3.2-Think requires \textbf{1313.3 seconds}. This demonstrates that for strict physical execution, combining an agile instruction-following model with a rigid symbolic controller is a far more pragmatic solution than blindly scaling reasoning compute.

\begin{table*}[t!]
\centering
\scriptsize
\renewcommand{\arraystretch}{1.3} 
\caption{\textbf{Side-by-Side Execution Log Comparison.} BioProAgent utilizes FSM states to progress deterministically, completing the task in 8 steps. In contrast, AutoGPT enters a semantic drift loop, exhausting the context window (Step 45) without producing valid code.}
\label{tab:log_comparison}
\begin{tabular}{p{0.05\textwidth} | p{0.45\textwidth} | p{0.42\textwidth}}
\toprule
\textbf{Step} & \textbf{BioProAgent (Ours)} & \textbf{AutoGPT (Baseline)} \\ \midrule

\textbf{1} & 
\textbf{[State: \texttt{CLARIFY\_INTENT}]} \newline
\textit{Trigger:} $\sigma_{ambiguous}$ \newline
\textit{Action:} \texttt{clarify\_experiment\_scope(query=...)} \newline
\textit{Output:} "Sample Volume: 50uL, Kit ID: 10038" & 
\textbf{[Thought]} I need to find the protocol for metabolomics. \newline
\textit{Action:} \texttt{retrieve\_knowledge(query="metabolomics prep")} \\ \hline

\textbf{2} & 
\textbf{[State: \texttt{DESIGN\_DRAFT}]} \newline
\textit{Trigger:} $\sigma_{know} \land \neg \sigma_{draft}$ \newline
\textit{Action:} \texttt{generate\_scientific\_draft(...)} \newline
\textit{Grounding ($\Phi$):} Mapped "Kit\_A" $\to$ \texttt{ID:10038} & 
\textbf{[Thought]} I found some docs. Now I need specific reagent list. \newline
\textit{Action:} \texttt{retrieve\_knowledge(query="reagent list")} \\ \hline

\textbf{3} & 
\textbf{[State: \texttt{VERIFY\_DRAFT}]} \newline
\textit{Trigger:} $\sigma_{draft} \land (\sigma_{sci} = 0)$ \newline
\textit{Action:} \texttt{reflect\_on\_protocol(...)} \newline
\textit{Result:} Scientific Logic Approved ($\sigma_{sci} \leftarrow 1$). & 
\textbf{[Thought]} I need to check the temperature settings. \newline
\textit{Action:} \texttt{retrieve\_knowledge(query="temperature settings")} \\ \hline

\textbf{...} & 
\textit{(Step 4-7: State transitions through DESIGN\_CODE and Physical Rule Verification $\mathcal{K}_p$)} & 
\textit{(Repeated retrieval loops for 30+ steps...)} \\ \hline

\textbf{8 / 45} & 
\textbf{[State: \texttt{SUCCESS}]} \newline
\textit{Trigger:} $\sigma_{code} \land (\sigma_{phys} = 1)$ \newline
\textit{Result:} Valid JSON Code committed. \newline
\textcolor{green!60!black}{\textbf{Task Success (Total Tokens: 10,329)}} & 
\textbf{[Error]} \textcolor{red}{\textbf{Max Context Exceeded.}} \newline
\textit{Result:} Timeout / Fail. \newline
\textcolor{red}{\textbf{Task Failure (Total Tokens: 59,278)}} \\ \bottomrule
\end{tabular}
\end{table*}

\section{Case Studies \& Logs}
\subsection{Detailed Execution Traces (Subset C)}
\label{app:trace_comparison}

To provide granular insight into the ``Efficiency Gap'' visualized in Figure~\ref{fig:trace_divergence}, we present raw execution logs from a representative long-horizon task (Task ID: \texttt{Metabolomics-Sample-Prep}). Table~\ref{tab:log_comparison} contrasts the decision-making trajectory of BioProAgent against AutoGPT. While BioProAgent leverages its deterministic FSM backbone ($s_{t+1} = \Delta(\sigma_t, s_t)$) to strictly enforce the \textsc{Design-Verify-Rectify} cadence, AutoGPT succumbs to a ``semantic drift loop,'' repeatedly querying redundant information without converging on an executable plan.

\subsection{Taxonomy of Baseline Failures}
\label{app:failure_taxonomy}

Expanding on the trace analysis in Figure~\ref{fig:trace_divergence}, we categorize the failure modes of baseline agents into two distinct pathologies based on the execution logs across our benchmark evaluations:

\paragraph{Context Paralysis (Dominant in ReAct).} 
As observed in the vast majority of ReAct failures (visualized in Figure~\ref{fig:trace_divergence}, Middle), the agent correctly retrieves the schema but fails to synthesize it into code. The logs show abrupt termination with empty outputs. This confirms that without \textbf{Symbol Grounding ($\Phi$)}, the raw HTML/JSON payloads from API docs quickly saturate the LLM's working memory, leading to a cognitive freeze.

\paragraph{Info-Seeking Loop (Dominant in AutoGPT).} 
In almost all instances of AutoGPT failures (visualized in Figure~\ref{fig:trace_divergence}, Bottom), the agent enters a recursive retrieval loop (e.g., querying buffer composition $\to$ ``buffer pH'' $\to$ ``buffer supplier''). Unlike BioProAgent's deterministic FSM, which mandates a rigid progression from \texttt{DESIGN\_DRAFT} through \texttt{VERIFY\_DRAFT} to \texttt{DESIGN\_CODE}, AutoGPT lacks a structural stopping condition, optimizing for \textit{information completeness} rather than \textit{task completion}.

\section{System Prompts}
\label{app:prompts}

To support reproducibility, we provide the core system prompts used in BioProAgent. Variable placeholders are denoted by brackets (e.g., \texttt{\{user\_input\}}).

\subsection{FSM-Driven Planner Prompt}
\label{app:prompt_planner}
The planner prompt illustrates how the Finite State Machine (FSM) explicitly governs the agent's trajectory. Note the \texttt{State->Action Mapping} section, which enforces the \textbf{Priority Decision Matrix} described in Section~\ref{sec:planning} (Table 1), ensuring that the agent strictly adheres to the \textbf{Design-Verify-Rectify} workflow.

\begin{lstlisting}[caption={The Neuro-Symbolic Planner Prompt}, label={lst:planner}, basicstyle=\ttfamily\scriptsize, breaklines=true]
You are BioProAgent, a bio-experiment planning agent. Your task is to convert user requests into executable tool call sequences.

## Scientific Principles
- Reproducibility: Steps are clear and reproducible.
- Control Principle: Negative control (NTC), Positive control (PTC), Blank control.
- Variable Control: Change only one variable at a time.

## Current State: {current_state}
## Suggested Actions: {state_guidance}

## User Request
{user_input}

## Experiment Context
{experiment_context}

## Executed Steps
{execution_summary}

## Available Tools
{tools_description}

## Output Format Requirements (Strictly Follow)
Output must be a JSON array. 
Example: [{"tool_name": "retrieve_knowledge", "args": {...}}]

### State -> Action Mapping (Hard Constraints)
- CLARIFY_INTENT -> [clarify_experiment_scope]
- DESIGN_DRAFT   -> [retrieve_knowledge, generate_scientific_draft]
- VERIFY_DRAFT   -> [reflect_on_protocol]  
- RECTIFY_DRAFT  -> [modify_protocol, reflect_on_protocol]
- DESIGN_CODE    -> [align_draft_to_automation, generate_machine_code]
- RECTIFY_CODE   -> [fix_machine_code, validate_machine_code]
- SUCCESS        -> [ask_user_confirmation]

## Important Instructions
- Follow scientific principles strictly.
- Current state is **{current_state}**
- Suggested actions: {state_guidance}
- The FSM state determines your valid action space. Do not hallucinate tools outside the mapping.
\end{lstlisting}

\subsection{Neuro-Symbolic Alignment Prompt}
\label{app:prompt_alignment}
This prompt operationalizes the \textbf{Semantic Symbol Grounding ($\Phi$)} phase. It maps natural language protocols to the hardware registry by categorizing steps into \texttt{[AUTO]}, \texttt{[EXTERNAL]}, and \texttt{[MANUAL]}, and strictly grounding natural language reagents to valid registry IDs to prevent context overflow.

\begin{lstlisting}[caption={Protocol Alignment and Device Grounding Prompt}, label={lst:alignment}, basicstyle=\ttfamily\scriptsize, breaklines=true]
You are a biological experiment automation expert.
Task: Convert the [Reference Info] into a standardized executable process.

## Step Tagging Rules (Critical!)
Every step must be tagged with one of the following:

| Tag | Meaning | Usage |
|-----|---------|-------|
| **[AUTO]** | Executable by platform | Use available devices below |
| **[EXTERNAL]** | Requires external device | Lab has device, but not integrated |
| **[MANUAL]** | Requires human operation | Decision making or special ops |

**Important**: [EXTERNAL] and [MANUAL] are NOT failures. They are normal parts of complex biological workflows.

## Symbol Grounding ($\Phi$) Rules
When assigning an [AUTO] step, you MUST map fuzzy reagent/container names to strict semantic pointers (e.g., map "Kit_A" -> "ID:10038") based on the Provided Context. Do NOT pass full payload schemas into the sequence.

## Available Automation Devices (Summary)
- Pipetting Workstation: Liquid transfer (0.5uL-1mL), mixing, temp control
- PCR Machine: Temp cycling (4-99C)
- Sealer/Peeler: Plate sealing
- Centrifuge: Separation (500-15000g)
- Incubator: Constant temp (20-60C)
- Plate Reader: Absorbance/Fluorescence

## Output Format Specification
1. [AUTO] Pipetting Workstation - Setup PCR Reaction
   1.1 [AUTO] Pipetting Workstation - Add Template DNA (Vol=2uL)
     - Source: [Grounded_ID: template_tube_1]
     - Target: [Grounded_ID: pcr_plate_01]
   1.2 [AUTO] Pipetting Workstation - Add Master Mix
2. [AUTO] Sealer - Seal Plate (Temp=170C)
3. [AUTO] PCR Machine - Amplification
   - Program: 95C 5min -> [95C 30s, 58C 30s, 72C 1min] x35
4. [EXTERNAL] Gel Electrophoresis - Check Products
   - Note: Requires external imaging system
5. [MANUAL] Human Decision - Verify Band Size
   - Criteria: Target band at ~500bp

## Input Information
User Requirements: {exp_info}
Reference Protocol: {protocol}
\end{lstlisting}

\subsection{Baseline SOP Injection}
\label{app:prompt_baseline}
To ensure fair comparison, baselines (ReAct, Direct LLM) were injected with a Standard Operating Procedure (SOP) prompt. This minimizes naive formatting errors and ensures that performance gaps are attributable to reasoning architecture rather than prompt engineering.

\begin{lstlisting}[caption={Standard Operating Procedure (SOP) for Baselines}, label={lst:baseline}, basicstyle=\ttfamily\scriptsize, breaklines=true]
## Laboratory Standard Operating Procedures (SOP)
You are a Senior Laboratory Automation Engineer. To ensure success, you MUST follow these protocols:

1. **Tool Priority**: Do NOT hallucinate JSON formats. 
   - Use `generate_machine_code` to create the final JSON. 
   - This tool knows the correct Schema (nodes, resourceId, connections).
   
2. **Validation Loop**:
   - After generating code, you MUST use `validate_machine_code`.
   - If validation fails (red light), use `fix_machine_code` or regenerate.
   - Do NOT output the final answer until validation passes.

3. **Data Handling**: 
   - Use data references (e.g., $draft, $machine_code) to pass information between tools to avoid context overflow.
\end{lstlisting}

\end{document}